\documentclass[12pt]{article}

\usepackage[a4paper,margin=2.5cm]{geometry}
\usepackage[T1]{fontenc}
\usepackage[utf8]{inputenc}
\usepackage[english]{babel}
\usepackage{lmodern}
\usepackage{microtype}

\usepackage{authblk}
\usepackage{amsmath,amssymb}
\usepackage{graphicx}
\usepackage{subcaption}
\usepackage{booktabs}
\usepackage{hyperref}

\usepackage{array}
\usepackage{tabularx}

\usepackage{algorithm}
\usepackage{algpseudocode}

\usepackage{pifont}
\newcommand{\cmark}{\ding{51}}
\newcommand{\xmark}{\ding{55}}

\newcolumntype{P}[1]{>{\raggedright\arraybackslash}p{#1}}
\newcolumntype{Y}{>{\raggedright\arraybackslash}X}
\newcolumntype{C}{>{\centering\arraybackslash}X}

\title{Calendar-Structured Sparse Principal Component Analysis for
Interpretable Multi-Periodic Electricity Consumption Profiles}

\author[1]{Carlos Quesada-Granja\thanks{Corresponding author. \texttt{c.quesada@ua.es}}}
\author[2]{Tony Castillo-Calzadilla}
\author[3]{Carlos Rizo-Maestre}

\affil[1]{\small
Dept.\ of Computer Science and Artificial Intelligence,
University of Alicante.
Ctra.\ de San Vicente del Raspeig s/n,
03690 San Vicente del Raspeig, Alicante, Spain.
}

\affil[2]{\small
Dept.\ of Industrial Engineering,
University of La Laguna.
Camino San Francisco de Paula s/n,
38200 San Cristóbal de La Laguna, Santa Cruz de Tenerife, Spain.
}

\affil[3]{\small
Dept.\ of Architectural Constructions,
University of Alicante.
Ctra.\ de San Vicente del Raspeig s/n,
03690 San Vicente del Raspeig, Alicante, Spain.
}

\date{}

\begin{document}

\maketitle

\begin{abstract}
Long-term electricity-consumption profiles exhibit several simultaneous
periodic structures, including daily, weekly, and annual cycles. This work
introduces Calendar-Structured Sparse Principal Component Analysis
(Calendar-SPCA), a structured representation-learning method that incorporates
this known multi-periodic geometry directly into a low-dimensional
factorization. The method represents the feature domain as the Cartesian
product of cyclic calendar axes and combines an $\ell_1$ loading penalty with
graph total variation, producing sparse, locally coherent, and directly
interpretable latent factors. In this study, Calendar-SPCA is applied to the
interpretable analysis of long-term electricity-consumption profiles and
evaluated on two independent public smart-meter datasets, GoiEner and Low
Carbon London, with different population sizes and temporal resolutions. A
factorial experiment characterizes the effects of sparsity and calendar
coherence and examines robustness across sample size, latent dimensionality,
and repeated fits. At rank 15, Calendar-SPCA retains 96.92\% and 82.90\% of
the explained variance of rank-matched principal component analysis (PCA) in
GoiEner and Low Carbon London, respectively, with mean loading sparsities of
61.95\% and 81.50\%. Comparisons with classical sparse PCA and sparse PCA with
total variation (SPCA-TV) show that Calendar-SPCA organizes latent factors into
interpretable structures over the daily, weekly, and annual calendar axes
while preserving substantial low-rank information.
\end{abstract}



\section{Introduction}
\label{sec:introduction}

Electricity-consumption time series exhibit three well-established calendar
seasonalities: daily, weekly, and annual
\cite{quesada2024electricity}. Daily and weekly repetition is readily visible
over relatively short observation windows, while records spanning complete
years additionally expose the annual cycle. The growing availability of
long-term, high-resolution smart-meter data therefore makes it possible to
represent these three periodic structures simultaneously at the level of
individual consumers.

\begin{figure*}[!htbp]
    \centering
    \begin{minipage}[c]{0.49\textwidth}
        \centering
        \includegraphics[width=\linewidth]{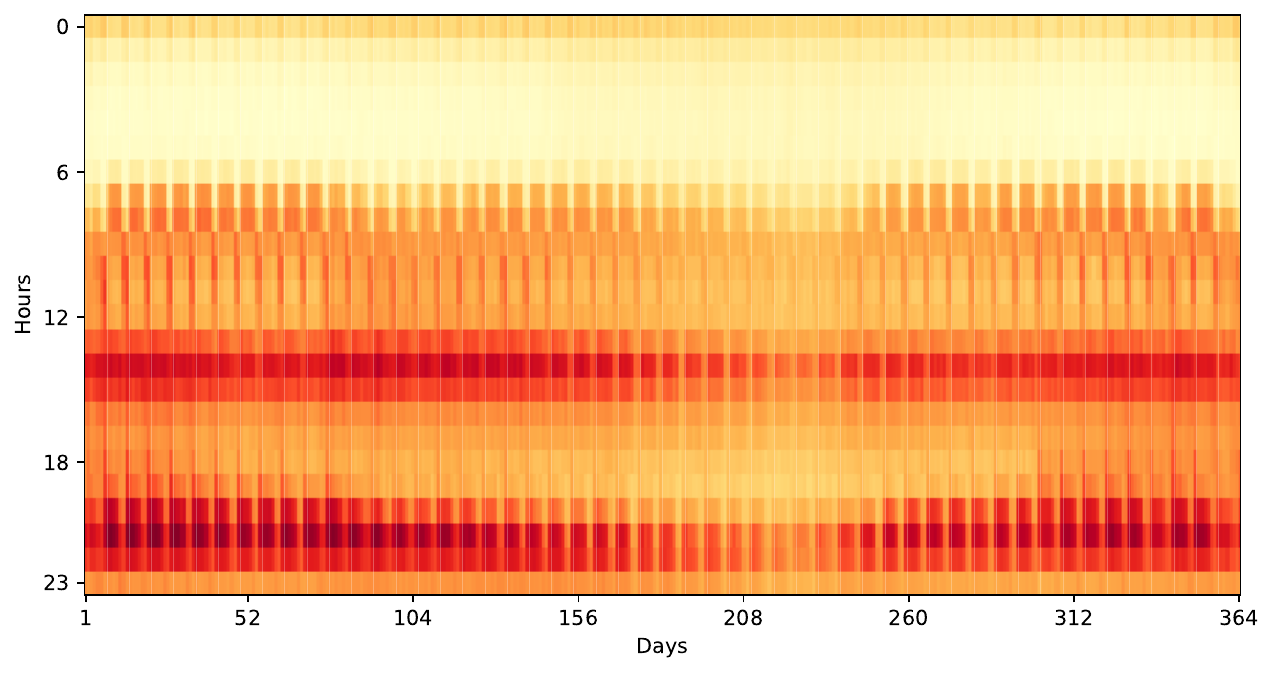}
    \end{minipage}
    \hfill
    \begin{minipage}[c]{0.49\textwidth}
        \centering
        \includegraphics[width=\linewidth]{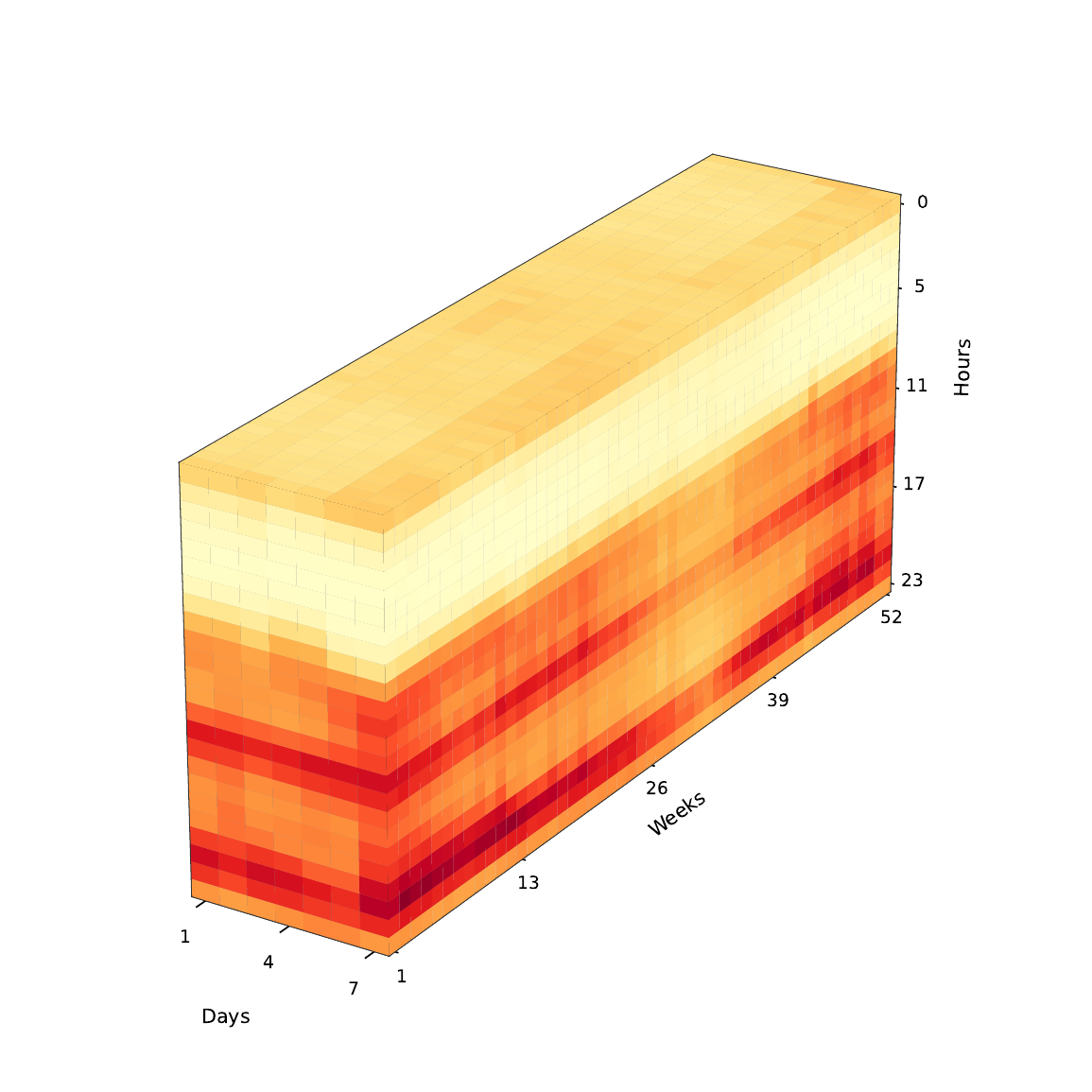}
    \end{minipage}
    \caption{
    Two representations of the same illustrative annual
    electricity-consumption pattern.
    Left: conventional heatmap representation over 24 hourly positions and 364 consecutive days. Right: calendar representation as a $24\times7\times52$ tensor, with
    axes corresponding to hour of day, day of week, and week of year.
    Calendar-SPCA uses this organization to define the geometry of the feature
    space.
    }
    \label{fig:calendar_representation}
\end{figure*}

Each measurement then occupies a known position within three periodic
coordinates: time of day, day of week, and position within the annual cycle.
Figure~\ref{fig:calendar_representation} illustrates this organization for an
hourly annual electricity-consumption profile. Such a profile can be displayed
as a conventional $24\times364$ heatmap. Here, we propose instead arranging
the same annual profile as a $24\times7\times52$ calendar tensor, whose axes
correspond directly to hour of day, day of week, and week of year. This
representation makes the multi-periodic geometry of the feature space
explicit.

The three calendar axes are cyclic. Hours connect across midnight, days connect
across the weekly boundary, and weeks connect across the annual boundary.
An annual profile can therefore be regarded as a signal defined over a
three-dimensional periodic domain. This observation provides a direct way to
incorporate calendar knowledge into dimensionality reduction.

Calendar-SPCA represents this domain through the Cartesian product of cycle
graphs,

\begin{equation}
    G =
    C_h
    \mathbin{\square}
    C_d
    \mathbin{\square}
    C_w,
    \label{eq:intro_calendar_graph}
\end{equation}

where $h$, $d$, and $w$ denote the numbers of positions along the daily,
weekly, and annual axes. The resulting graph connects neighbouring calendar
positions along each periodic coordinate and provides an explicit topology for
the loading space.

Given a centered data matrix $X_c$, Calendar-SPCA learns a low-rank
representation $X_c \approx UV^\top$ by solving

\begin{equation}
\min_{U,V}
\left\{
\underbrace{
\frac{1}{2}
\left\|X_c-UV^\top\right\|_F^2
}_{\text{reconstruction}}
+
\underbrace{
\vphantom{
\frac{1}{2}
\left\|X_c-UV^\top\right\|_F^2
}
\lambda_1\|V\|_{1,1}
}_{\text{loading sparsity}}
+
\underbrace{
\vphantom{
\frac{1}{2}
\left\|X_c-UV^\top\right\|_F^2
}
\lambda_{\mathrm{TV}}
\|D_GV\|_{1,1}
}_{\text{calendar coherence}}
\right\},
\label{eq:intro_calendar_spca}
\end{equation}

where $U$ contains the component scores, $V$ contains the loading vectors, and
$D_G$ is the incidence matrix of the calendar graph
(see Section~\ref{sec:calendar_factorization} for the complete formulation).
The reconstruction term
captures the dominant low-rank structure of the population, the
$\ell_1$ penalty concentrates each loading on a selected set of calendar
positions, and the graph total-variation penalty promotes coherent values
across neighbouring positions. The recovered components are therefore
expressed directly as sparse, structured patterns over the calendar domain.

This representation gives each loading an immediate temporal interpretation.
Each loading can be viewed as a sparse, coherent mask over the calendar
tensor, selecting the temporal regions associated with one latent mode of
variation. Active regions can describe particular times of day, weekdays,
seasons, or combinations of these coordinates, while the cyclic topology
preserves continuity across the boundaries of each periodic axis.
Calendar-SPCA thus connects dimensionality reduction with the
calendar coordinates naturally used to describe long-term electricity
consumption.

The contributions of this work are threefold:

\begin{itemize}

    \item \textbf{Multi-periodic calendar representation.}
    We formulate the feature space of long-term electricity profiles as the
    Cartesian product of cyclic temporal axes, explicitly encoding daily,
    weekly, and annual adjacency.

    \item \textbf{Calendar-structured sparse factorization.}
    We combine sparse loading estimation with graph total variation over the
    calendar topology to learn latent factors that are selective, locally
    coherent, and directly interpretable in temporal coordinates.

    \item \textbf{Quantitative and visual validation on real smart-meter data.}
    We evaluate Calendar-SPCA on two independent electricity-consumption
    datasets with different populations and temporal resolutions, characterize
    its regularization and robustness properties, and compare the resulting
    representations with PCA, sparse PCA, and SPCA-TV.

\end{itemize}

The resulting framework provides a compact representation of long-term
electricity profiles in which the known geometry of the calendar becomes part
of the representation-learning model itself.


\section{Related Work and Positioning}
\label{sec:related_work}

Calendar-SPCA connects several established lines of dimensionality-reduction
research: principal component analysis, sparse loading estimation, structured
regularization, total variation, and multiway representations.

Principal component analysis (PCA) represents high-dimensional observations
through a reduced number of latent directions that capture dominant modes of
variation \cite{jolliffe2016principal}. Sparse PCA extends this framework by
concentrating each loading on a selected subset of variables, strengthening
the link between latent components and the original feature space.
Representative formulations include SCoTLASS
\cite{jolliffe2003modified}, elastic-net sparse PCA
\cite{zou2006sparse}, regularized low-rank approximation
\cite{shen2008sparse}, and penalized matrix decomposition
\cite{witten2009penalized}.

Structured sparse methods incorporate relationships among variables into the
loading estimation. Group penalties encode predefined support structures
\cite{jenatton2010structured}, while fusion penalties promote similar
coefficients across related variables. The fused lasso
\cite{tibshirani2005sparsity} and generalized lasso
\cite{tibshirani2011solution} provide general formulations for this principle,
which has also been incorporated into sparse component models
\cite{guo2010principal,li2017incorporating}.

Total-variation regularization provides a particularly close connection to
Calendar-SPCA. SPCA-TV combines sparse component estimation with total
variation to obtain localized and coherent loading patterns over structured
feature domains \cite{depierrefeu2018spcatv}. Its components are extracted
sequentially through rank-one approximation and deflation.
Calendar-SPCA incorporates calendar regularization within a joint $K$-factor
reconstruction defined over a cyclic Cartesian-product graph, so that all
latent calendar factors are estimated simultaneously within the same
representation.

Tensor and multiway methods provide a complementary perspective by organizing
data along explicit modes such as consumers, hours, weekdays, or seasonal
positions \cite{kolda2009tensor,banin2025predicting}. Their multi-axis
representation highlights the value of preserving known structure across
several dimensions. In Calendar-SPCA, these axes jointly define the geometry
of the loading domain through their Cartesian product.

Table~\ref{tab:method_positioning} summarizes the resulting methodological
positioning in terms of five structural properties: sparse loadings, local
coherence, prior feature structure, cyclic axes, and joint factor estimation.

\begin{table*}[!ht]
\centering
\caption{
Positioning of Calendar-SPCA with respect to representative
dimensionality-reduction families. The entries summarize the principal
structural mechanisms emphasized by each family; individual methods within a
family may differ in their precise optimization and factorization strategy.
}
\label{tab:method_positioning}

\small
\renewcommand{\arraystretch}{1.35}
\setlength{\tabcolsep}{3pt}
\renewcommand{\tabularxcolumn}[1]{m{#1}}

\begin{tabularx}{\textwidth}{
    >{\raggedright\arraybackslash}m{0.23\textwidth}
    CCCCC
}
\hline

\textbf{Method family}
&
\shortstack{\textbf{Sparse}\\\textbf{loadings}}
&
\shortstack{\textbf{Local}\\\textbf{coherence}}
&
\shortstack{\textbf{Prior}\\\textbf{structure}}
&
\shortstack{\textbf{Cyclic}\\\textbf{axes}}
&
\shortstack{\textbf{Joint}\\\textbf{estimation}}
\\

\hline

PCA
& \xmark & \xmark & \xmark & \xmark & \cmark
\\

Sparse PCA
& \cmark & \xmark & \xmark & \xmark & method-dependent
\\

Structured / fused SPCA
& \cmark & \cmark & \cmark & method-dependent & method-dependent
\\

SPCA-TV
& \cmark & \cmark & TV operator & operator-dependent & sequential
\\

Tensor / multiway decompositions
& method-dependent & method-dependent & multi-axis & method-dependent & \cmark
\\

\textbf{Calendar-SPCA}
& \cmark & \cmark & \cmark & \cmark & \cmark
\\

\hline
\end{tabularx}
\end{table*}

Calendar-SPCA combines these properties within a single representation:
loading sparsity provides feature selectivity, graph total variation promotes
local coherence, the calendar supplies the prior feature structure, its
temporal coordinates form multiple cyclic axes, and the latent factors are
estimated jointly within a common low-rank factorization.


\section{Calendar-SPCA}
\label{sec:calendar_spca}

Calendar-SPCA represents long-term electricity profiles on their natural
multi-periodic calendar domain and incorporates this geometry directly into a
sparse low-rank factorization. The method consists of three elements: a cyclic
calendar graph defining relationships among features, a structured sparse
factorization over this graph, and an alternating optimization procedure for
estimating the latent factors.

\subsection{Calendar geometry}
\label{sec:calendar_graph}

Let

\[
X\in\mathbb{R}^{N\times M}
\]

contain $N$ electricity-consumption profiles, with each column corresponding
to one position of a common calendar representation. For three temporal axes
of sizes $h$, $d$, and $w$, the feature domain contains

\[
M=hdw
\]

positions indexed by

\[
(i_h,i_d,i_w),
\qquad
i_h\in\{0,\ldots,h-1\},
\quad
i_d\in\{0,\ldots,d-1\},
\quad
i_w\in\{0,\ldots,w-1\}.
\]

In the applications considered here, these coordinates represent time within
the day, day of week, and week of year. Hourly profiles therefore use a
$24\times7\times52$ domain, while half-hourly profiles use a
$48\times7\times52$ domain.

Calendar-SPCA defines adjacency on this domain through the Cartesian product

\begin{equation}
    G =
    C_h
    \mathbin{\square}
    C_d
    \mathbin{\square}
    C_w,
    \label{eq:calendar_graph}
\end{equation}

where $C_m$ denotes a cycle graph with $m$ vertices. Two calendar positions are
adjacent when they differ by one step along one temporal coordinate and share
the same values along the other two coordinates.

Each factor graph is cyclic, so adjacency also connects the boundaries of its
corresponding temporal axis. The resulting graph therefore preserves
continuity across midnight, across the weekly transition, and across the
annual boundary. Every calendar position is connected locally along the daily,
weekly, and annual directions, producing a single multi-periodic geometry for
the complete loading domain.

Let

\[
D_G\in\mathbb{R}^{|E|\times M}
\]

denote an oriented incidence matrix of $G$. For a loading vector
$v\in\mathbb{R}^{M}$, the vector $D_Gv$ contains differences between loading
coefficients at adjacent calendar positions. Its graph total variation is

\begin{equation}
    \|D_Gv\|_1
    =
    \sum_{(i,j)\in E(G)}
    |v_i-v_j|.
    \label{eq:graph_tv}
\end{equation}

This quantity gives a direct measure of local variation over the calendar.
Large differences contribute strongly when neighbouring calendar positions
receive different loading values, while coherent regions contribute little
when adjacent positions carry similar values. Because the edges of $G$ span
all three cyclic axes, the same measure captures coherence simultaneously
within the day, across the week, and throughout the annual cycle.

For a loading matrix
$V=[v_1,\ldots,v_K]\in\mathbb{R}^{M\times K}$, Calendar-SPCA uses the aggregate
graph total variation

\begin{equation}
    \|D_GV\|_{1,1}
    =
    \sum_{k=1}^{K}\|D_Gv_k\|_1,
    \label{eq:graph_tv_matrix}
\end{equation}

which provides the structural regularizer for the factorization introduced
next.

\subsection{Structured sparse factorization}
\label{sec:calendar_factorization}

Calendar-SPCA operates on the centered data matrix

\begin{equation}
    X_c
    =
    X-\mathbf{1}\mu^\top,
    \qquad
    \mu
    =
    \frac{1}{N}X^\top\mathbf{1},
    \label{eq:centered_data}
\end{equation}

where $\mu\in\mathbb{R}^{M}$ is the empirical mean profile. The centered
observations are represented through a $K$-component factorization

\begin{equation}
    X_c
    \approx
    UV^\top,
    \label{eq:calendar_factorization}
\end{equation}

with
$U=[u_1,\ldots,u_K]\in\mathbb{R}^{N\times K}$ containing the component scores
and
$V=[v_1,\ldots,v_K]\in\mathbb{R}^{M\times K}$ containing the corresponding
calendar loadings.

Calendar-SPCA estimates both factors through

\begin{equation}
\min_{U,V}
\left\{
\frac{1}{2}
\left\|X_c-UV^\top\right\|_F^2
+
\lambda_1\|V\|_{1,1}
+
\lambda_{\mathrm{TV}}\|D_GV\|_{1,1}
\right\},
\label{eq:calendar_spca_objective}
\end{equation}

subject to unit-norm constraints on the active score vectors,

\begin{equation}
    \|u_k\|_2=1,
    \qquad
    k\in\mathcal{A},
    \label{eq:score_normalization}
\end{equation}

where $\mathcal{A}$ denotes the set of active components.

The three terms in Eq.~\eqref{eq:calendar_spca_objective} define the main
properties of the representation. The reconstruction term captures dominant
variation in the centered profiles. The entrywise $\ell_1$ penalty promotes
selective loading supports, with $\lambda_1$ controlling the degree of
sparsity. The graph total-variation penalty promotes coherent loading values
across neighbouring calendar positions, with $\lambda_{\mathrm{TV}}$
controlling the strength of calendar regularization.

Their joint action gives each component a sparse and structured temporal
support. The $\ell_1$ term concentrates the loading on selected calendar
positions, while graph total variation encourages those positions to form
coherent patterns along the daily, weekly, and annual directions. The learned
loading vectors can therefore be visualized directly on the calendar domain
introduced in Section~\ref{sec:calendar_graph}.

The normalization in Eq.~\eqref{eq:score_normalization} resolves the scale
indeterminacy between each score vector $u_k$ and its loading $v_k$. Under
this convention, the loading magnitude carries the strength of the component,
while the normalized score vector describes its expression across the
observations.

The components are estimated jointly within the same low-rank factorization
and may represent correlated modes of variation. This flexibility allows
several complementary calendar structures to coexist within the learned
representation.

The nominal rank $K$ specifies the requested number of components. Components
that sustain a non-degenerate solution form the active set $\mathcal{A}$,
yielding the effective rank

\begin{equation}
    K_{\mathrm{eff}}
    =
    |\mathcal{A}|
    \leq K.
    \label{eq:effective_rank}
\end{equation}

The resulting model is thus defined by a compact set of jointly estimated
calendar loadings and their associated score vectors.

\subsection{Optimization}
\label{sec:optimization}

Calendar-SPCA is fitted by alternating between updates of the score matrix
$U$ and the loading matrix $V$. This block-coordinate structure preserves the
joint low-rank factorization while allowing each subproblem to exploit its
specific form.

For fixed $V$, the score vectors are updated conditionally on the remaining
components. For an active component $k$,

\begin{equation}
    \widetilde{u}_k
    =
    X_c v_k
    -
    \sum_{j\neq k}
    u_j
    \left(v_j^\top v_k\right),
    \label{eq:score_update}
\end{equation}

followed by

\begin{equation}
    u_k
    =
    \frac{\widetilde{u}_k}
         {\|\widetilde{u}_k\|_2}.
    \label{eq:score_update_normalized}
\end{equation}

Sweeps over the active components refine the score matrix while preserving the
unit-norm convention introduced in
Eq.~\eqref{eq:score_normalization}.

For fixed $U$, the loading matrix is obtained from the convex problem

\begin{equation}
\min_V
\left\{
\frac{1}{2}
\|X_c-UV^\top\|_F^2
+
\lambda_1\|V\|_{1,1}
+
\lambda_{\mathrm{TV}}
\|D_GV\|_{1,1}
\right\}.
\label{eq:loading_subproblem}
\end{equation}

The smooth reconstruction term has gradient

\begin{equation}
    \nabla f(V)
    =
    V(U^\top U)-X_c^\top U.
    \label{eq:loading_gradient}
\end{equation}

The sparse and graph-total-variation terms are handled with a Condat--V\~u
primal--dual splitting scheme
\cite{condat2013primal,vu2013splitting}. Soft-thresholding acts on the loading
variables associated with the $\ell_1$ penalty, while dual variables associated
with $D_G$ enforce the calendar total-variation regularization. The explicit
primal--dual updates and step-size construction are given in
Appendices~\ref{app:loading_update} and~\ref{app:step_sizes}, respectively.

The complete fitting procedure therefore alternates between score refinement
and solution of the structured loading subproblem until the low-rank
representation stabilizes. Convergence is assessed from the stabilization of
the objective and reconstructed representation together with convergence of
the loading subproblem, as detailed in
Appendix~\ref{app:convergence}.

The optimization also maintains the active set $\mathcal{A}$ and applies
component recovery from residual variation when a component becomes
degenerate. Components that retain a stable solution form the effective rank
$K_{\mathrm{eff}}$. The initialization and component-recovery procedures are
described in Appendix~\ref{app:initialization_recovery}, and
Algorithm~\ref{alg:calendar_spca} summarizes the main fitting procedure.

\begin{algorithm}[!ht]
\caption{Calendar-SPCA fitting procedure}
\label{alg:calendar_spca}
\begin{algorithmic}[1]
\Require Centered matrix $X_c$, nominal rank $K$,
         calendar incidence matrix $D_G$,
         $\lambda_1$, $\lambda_{\mathrm{TV}}$
\State Initialize $U$ and $V$ from a low-rank decomposition of $X_c$
\State Set $\mathcal{A}\gets\{1,\ldots,K\}$
\Repeat
    \For{$k\in\mathcal{A}$}
        \State Update $u_k$ using Eq.~\eqref{eq:score_update}
        \State Normalize $u_k$ using
        Eq.~\eqref{eq:score_update_normalized}
    \EndFor
    \State Update $V$ by solving
    Eq.~\eqref{eq:loading_subproblem} with primal--dual splitting
    \State Apply component recovery and update $\mathcal{A}$ when required
\Until{objective and reconstruction convergence criteria are satisfied}
\State \Return $U$, $V$, and $\mathcal{A}$
\end{algorithmic}
\end{algorithm}

The numerical settings used for score refinement, primal--dual optimization,
convergence, initialization, and component recovery are collected in
Appendix~\ref{app:numerical_settings}.


\section{Experimental Design}
\label{sec:experiments}

\subsection{Datasets}
\label{sec:datasets}

Calendar-SPCA was evaluated on two independent smart-meter datasets:
GoiEner, from electricity consumers in Spain, and Low Carbon London (LCL),
from residential consumers in the United Kingdom. The datasets provide
complementary evaluation settings through different population sizes and
temporal resolutions, while sharing the same daily, weekly, and annual
calendar organization.

Table~\ref{tab:datasets} summarizes the annual profiles used in the
experiments.

\begin{table*}[!ht]
\centering
\caption{
Smart-meter datasets used in the evaluation of Calendar-SPCA.
Both are represented as complete 52-week annual profiles organized along
daily, weekly, and annual calendar axes.
}
\label{tab:datasets}
\small
\begin{tabular}{lcc}
\hline
\textbf{Characteristic}
&
\textbf{GoiEner}
&
\textbf{Low Carbon London}
\\
\hline

Country
& Spain
& United Kingdom
\\

Annual profile years
& 2018--2023
& 2012--2013
\\

Temporal resolution
& 1 hour
& 30 minutes
\\

Calendar domain
& $24\times7\times52$
& $48\times7\times52$
\\

Features per profile ($M$)
& 8,736
& 17,472
\\

Annual profiles ($N$)
& 87,187
& 5,359
\\

Quality criterion
& $\leq72$ imputed samples
& $\leq144$ imputed samples
\\

\hline
\end{tabular}
\end{table*}

\paragraph{GoiEner.}

The GoiEner dataset contains long-term smart-meter electricity measurements
from supply points distributed throughout peninsular Spain. The data source,
anonymization procedure, and measurement processing are described in
\cite{quesada2024electricity}. The exact dataset release used in this study
is publicly available through Zenodo \cite{quesada2026goiener}.
Annual observations were constructed following the ISO-week procedure of
\cite{quesada2025data}, producing complete hourly profiles on a
$24\times7\times52$ calendar domain. For ISO years containing 53 weeks, the
first 52 complete weeks were retained to preserve a common feature domain.
Application of the profile-quality
criterion yielded $87{,}187$ annual profiles with $8{,}736$ features each.

\paragraph{Low Carbon London.}

The Low Carbon London dataset contains half-hourly residential electricity
measurements collected as part of the UK Power Networks Low Carbon London
project \cite{ukpowernetworks_lcl}. Source timestamps were converted from UTC/GMT
end-of-interval labels to Europe/London interval-start civil time and
regularized across daylight-saving transitions before construction of the
local ISO-calendar representation. Complete annual profiles were then
organized on a $48\times7\times52$ domain, providing twice the within-day
temporal resolution of GoiEner. Among the 5,374 profiles satisfying the
full-year and imputation criteria, 15 with a degenerate normalization scale
were excluded, yielding the final evaluation matrix of 5,359 profiles with
17,472 features each.

\bigskip
\noindent
For both datasets, missing measurements were imputed before annual
profile construction and the quality criteria in
Table~\ref{tab:datasets} were applied. Each retained profile $x$ was then
normalized using the robust scale

\begin{equation}
    s
    =
    \min\left\{
    Q_3+1.5(Q_3-Q_1),
    \max_t x_t
    \right\},
    \label{eq:profile_scale}
\end{equation}

where $Q_1$ and $Q_3$ are the first and third quartiles. This scale combines
the observed profile maximum with Tukey's upper inner fence
\cite{tukey1977exploratory}.

The normalized profile was obtained as

\begin{equation}
y_t
=
\begin{cases}
x_t/s,
& x_t\leq s,\\[2mm]
1+\alpha\log(x_t/s),
& x_t>s,
\end{cases}
\qquad
\alpha=0.2.
\label{eq:profile_normalization}
\end{equation}

The logarithmic continuation retains the relative magnitude of high-consumption
values while limiting their influence on the profile scale. This transformation
places annual profiles on a comparable scale while preserving their temporal
structure. Calendar-SPCA subsequently centers each dataset according to
Eq.~\eqref{eq:centered_data}.

\subsection{Experimental protocol}
\label{sec:experimental_protocol}

The experimental design first characterized the operating behaviour of
Calendar-SPCA across regularization strength, sample size, and latent
dimensionality. GoiEner was used for this broad analysis because its large
population supports controlled subsampling over a wide range of sample sizes.

The two regularization parameters were varied independently over

\begin{equation}
    \lambda_1,\lambda_{\mathrm{TV}}
    \in
    \{0,\;0.05,\;0.2,\;0.5,\;1,\;3,\;10\},
    \label{eq:regularization_grid}
\end{equation}

together with

\begin{equation}
    K\in\{5,\;10,\;15,\;20,\;25,\;30\}
\end{equation}

and

\begin{equation}
    N\in\{5{,}000,\;10{,}000,\;15{,}000,\;20{,}000\}.
\end{equation}

Five repetitions were fitted for every configuration, giving a total of
$5{,}880$ Calendar-SPCA fits. Across repetitions, both the sampled observations
and the initialization seed varied. Within each repetition, the four sample
sizes were nested, so larger subsets progressively extended the smaller ones.
Regularization settings sharing the same $N$, $K$, and repetition used the
same initialization seed.
This design supports paired comparisons across
the regularization grid and controlled analysis of the effects of increasing
sample size.

The factorial experiment serves two purposes in the main evaluation. First,
it reveals how the sparsity and graph-total-variation penalties shape
reconstruction quality and loading structure across the
$(\lambda_1,\lambda_{\mathrm{TV}})$ plane. Second, it characterizes the
robustness of the learned representation as the amount of data and the
requested rank vary.

Solution reproducibility was assessed by matching active loading vectors
across repeated fits of the same configuration using sign- and
permutation-invariant cosine similarity. The resulting repeat-stability metric is defined in Section~\ref{sec:evaluation_criteria}.

The factorial analysis was used to identify a common nominal rank for the
final representations by jointly considering reconstruction capacity,
loading structure, stability, and convergence across sample sizes. The
selected rank was subsequently used for both GoiEner and LCL in the
model-selection and cross-method experiments.

\subsection{Model selection and baselines}
\label{sec:model_selection}

With the common nominal rank selected from the factorial analysis, the final
regularization strength was selected independently for GoiEner and LCL through
a one-dimensional path

\begin{equation}
    \lambda_1=\lambda_{\mathrm{TV}}=\lambda.
    \label{eq:diagonal_regularization}
\end{equation}

This diagonal path varies both structural mechanisms jointly through a single
operating parameter. The pre-specified candidate path was

\[
\begin{aligned}
\lambda\in\{&
0,\;0.05,\;0.10,\;0.20,\;0.25,\;0.50,\;0.75,\;1.00,\;1.25,\\
&1.50,\;1.75,\;2.00,\;2.25,\;2.50,\;3,\;4,\;5,\;7,\;10
\}.
\end{aligned}
\]

For each candidate value of $\lambda$, reconstruction error was measured as

\begin{equation}
    \rho_\lambda
    =
    \left\|
    X_c-U_\lambda V_\lambda^\top
    \right\|_F,
    \label{eq:lcurve_residual}
\end{equation}

and the magnitude of the structured loading representation as

\begin{equation}
    \Omega_\lambda
    =
    \|V_\lambda\|_{1,1}
    +
    \|D_GV_\lambda\|_{1,1}.
    \label{eq:lcurve_structure}
\end{equation}

The resulting points

\begin{equation}
    \left(
    \log_{10}\Omega_\lambda,\,
    \log_{10}\rho_\lambda
    \right)
\end{equation}

form a discrete L-curve \cite{hansen1992analysis,hansen1993use}. 
Selection was restricted to converged full-rank solutions
($K_{\mathrm{eff}}=K$) along this path. 
Among the eligible interior points, 
the operating value was selected by maximum discrete Menger curvature.
Component visualizations and post-fit structural metrics were not used for
parameter selection.
The resulting dataset-specific operating points are reported in
Section~\ref{sec:results_lcurve}.

The final Calendar-SPCA representations were compared with three reference
methods at the same selected rank, spanning increasing levels of loading
structure. PCA provides the rank-matched linear reconstruction reference
\cite{jolliffe2016principal}. Sparse PCA was represented by the formulation
of Zou et al.~\cite{zou2006sparse}, providing sparse loading estimation.
SPCA-TV \cite{depierrefeu2018spcatv} adds total-variation regularization and
was supplied with the same cyclic calendar neighbourhood used by
Calendar-SPCA.

Regularization parameters for SPCA and SPCA-TV were selected numerically along
method-specific regularization paths before analysis of the resulting
component maps. The comparison therefore evaluates PCA, sparse loading
estimation, sparse total-variation components, and Calendar-SPCA under a common
rank and a common calendar-based structural assessment.

Implementation details for the reference methods and their regularization selection procedures are provided in Appendices~\ref{app:baseline_methods} and~\ref{app:baseline_selection}.

\subsection{Evaluation criteria}
\label{sec:evaluation_criteria}

The evaluation combines reconstruction quality, loading structure, and
solution reproducibility.

\paragraph{Explained variance.}

Reconstruction quality is measured through

\begin{equation}
    \operatorname{EV}
    =
    1-
    \frac{
        \|X_c-\widehat X\|_F^2
    }{
        \|X_c\|_F^2
    },
    \label{eq:explained_variance}
\end{equation}

where $\widehat X$ denotes the reconstruction used in the corresponding
analysis. In the Calendar-SPCA factorial experiments,
$\widehat X=UV^\top$ is the fitted reconstruction. For the cross-method
comparison, reconstruction quality is evaluated uniformly from the fitted
loading span; for non-orthogonal loading systems, $\widehat X$ is obtained by
least-squares projection onto that span.
For this comparison, explained variance is also expressed relative to
rank-matched PCA through

\begin{equation}
    \operatorname{Retention}_{\mathrm{PCA}}(m)
    =
    \frac{
        \operatorname{EV}_m
    }{
        \operatorname{EV}_{\mathrm{PCA}}
    }.
    \label{eq:pca_retention}
\end{equation}

This ratio measures the fraction of PCA explained variance retained by each
structured representation.

\paragraph{Component contribution.}

The importance of an individual Calendar-SPCA component is measured through
the increase in normalized reconstruction error produced by removing it from
the fitted representation,

\begin{equation}
    C_k
    =
    \frac{
        \|X_c-\widehat X_{-k}\|_F^2
        -
        \|X_c-UV^\top\|_F^2
    }{
        \|X_c\|_F^2
    },
    \label{eq:conditional_contribution}
\end{equation}

where
$\widehat X_{-k}=\sum_{j\neq k}u_jv_j^\top$.
Because the components are not required to be orthogonal, these conditional
contributions are not additive and need not sum to the total explained
variance. They are used here to indicate the relative importance of individual
components in the calendar visualizations.

\paragraph{Loading sparsity.}

For the numerical support

\[
    S_k
    =
    \{j:|v_{jk}|>\tau_s\},
    \qquad
    \tau_s=10^{-10},
\]

loading sparsity is defined as

\begin{equation}
    \operatorname{Sparsity}(v_k)
    =
    1-\frac{|S_k|}{M}.
    \label{eq:loading_sparsity}
\end{equation}

Higher values correspond to components concentrated on a smaller portion of
the calendar domain.

\paragraph{Calendar coherence.}

Local coherence is summarized through relative graph total variation,

\begin{equation}
    \operatorname{RTV}(v_k)
    =
    \frac{\|D_Gv_k\|_1}{\|v_k\|_1}.
    \label{eq:relative_tv}
\end{equation}

Lower RTV corresponds to more coherent loading values across neighbouring
daily, weekly, and annual positions. Reported model-level values are averaged
over the active components.

\paragraph{Calendar regions.}

The support $S_k$ induces connected regions on the cyclic calendar graph.
Effective regions are those containing at least $0.5\%$ of the active support
or at least $1\%$ of the loading's total $\ell_1$ mass. Their number summarizes
how compactly the active loading structure is organized over the calendar.

\paragraph{Stability.}

Reproducibility is evaluated by optimally matching active loading vectors
across fits using sign-invariant cosine similarity,

\begin{equation}
    c_{ij}
    =
    \frac{
        \left|
        {v_i^{(a)}}^\top v_j^{(b)}
        \right|
    }{
        \|v_i^{(a)}\|_2
        \|v_j^{(b)}\|_2
    }.
    \label{eq:loading_similarity}
\end{equation}

Model-level stability is then

\begin{equation}
    S
    =
    \frac{1}{|\mathcal{M}|}
    \sum_{(i,j)\in\mathcal{M}}
    c_{ij},
    \label{eq:stability_score}
\end{equation}

where $\mathcal{M}$ denotes the optimal one-to-one matching between active
components. Repeat stability measures agreement across repetitions of the same
configuration.

\paragraph{Numerical reliability.}

The effective rank $K_{\mathrm{eff}}$ is monitored throughout the evaluation.
Factorial summaries include configurations with at least four converged fits
among the five repetitions. Extended convergence and numerical diagnostics are
reported in Appendix~\ref{app:convergence_diagnostics}.


\section{Results}
\label{sec:results}

\subsection{What do the two regularizers do?}
\label{sec:results_regularization}

\begin{figure*}[!ht]
    \centering
    \includegraphics[width=\textwidth]{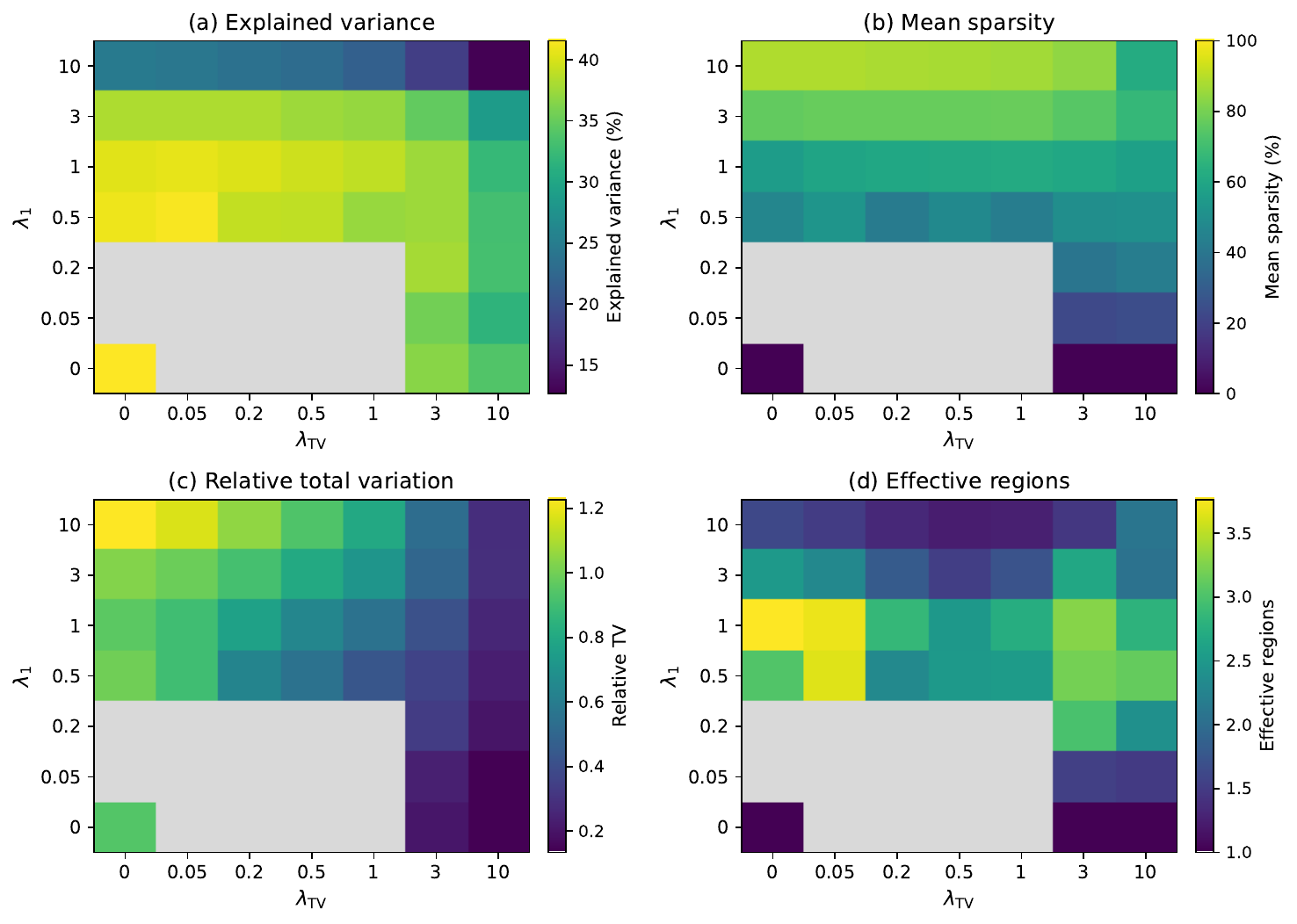}
    \caption{
    Regularization landscape of Calendar-SPCA on the GoiEner dataset.
    Heatmaps show
    (a) explained variance,
    (b) mean loading sparsity,
    (c) mean relative total variation, and
    (d) mean effective-region count
    across the $(\lambda_1,\lambda_{\mathrm{TV}})$ grid.
    Values are averaged across reliable $(N,K)$ scenarios.
    Grey cells indicate regularization pairs without a reliable factorial
    summary.
    }
    \label{fig:regularization_landscape}
\end{figure*}

The GoiEner factorial experiment described in
Section~\ref{sec:experimental_protocol} provides a direct view of how the two
Calendar-SPCA regularizers shape the learned representation.
Figure~\ref{fig:regularization_landscape} summarizes explained variance,
loading sparsity, relative total variation, and effective-region count across
the $(\lambda_1,\lambda_{\mathrm{TV}})$ plane.

The sparsity penalty $\lambda_1$ controls the amount of calendar support used
by the components. Increasing $\lambda_1$ progressively raises loading
sparsity, concentrating the representation on a smaller set of temporal
positions.

The graph total-variation penalty $\lambda_{\mathrm{TV}}$ controls local
calendar coherence. Increasing $\lambda_{\mathrm{TV}}$ reduces relative total
variation, promoting similar loading values across neighbouring positions of
the daily, weekly, and annual cycles.

The effective-region landscape captures the joint action of the two
regularizers on the organization of the active support into connected temporal
structures. Together, $\lambda_1$ and $\lambda_{\mathrm{TV}}$ provide
complementary control over selectivity and coherence, while explained variance
quantifies the associated reconstruction trade-off.

These results identify a broad operating region in which Calendar-SPCA
combines substantial reconstruction capacity with sparse and calendar-coherent
loading patterns. A matched four-configuration ablation further isolates the
contribution of each penalty in Appendix~\ref{app:penalty_ablation}. The next
analysis examines how consistently this behaviour is maintained across sample
size and nominal rank.

\subsection{Is the representation robust?}
\label{sec:results_robustness}

\begin{figure*}[!ht]
    \centering
    \includegraphics[
        width=\textwidth
    ]{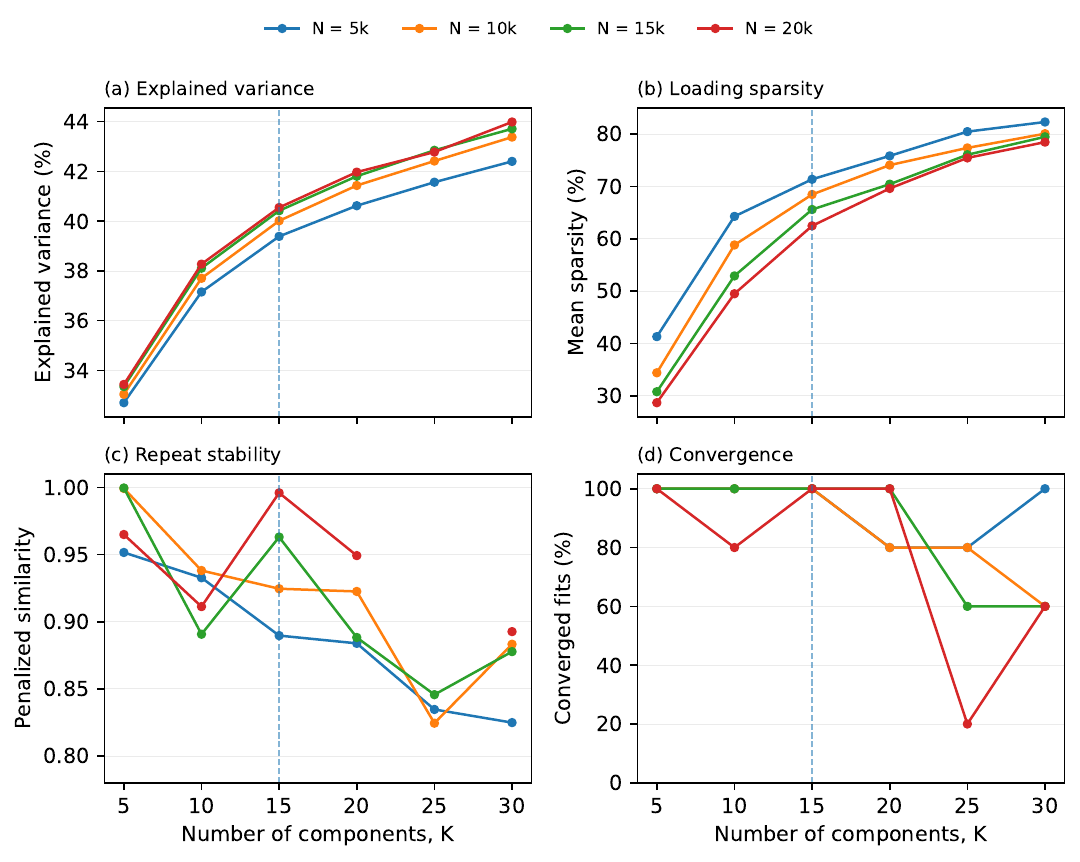}
    \caption{
    Robustness of Calendar-SPCA across sample size $N$ and nominal rank $K$ at
    $\lambda_1=\lambda_{\mathrm{TV}}=1$.
    (a) Mean explained variance,
    (b) mean loading sparsity,
    (c) repeat stability, and
    (d) convergence rate across five repeated fits.
    The dashed vertical line marks $K=15$, the common nominal rank used for the
    final representations.
    }
    \label{fig:robustness_sensitivity}
\end{figure*}

Robustness was examined at the common regularization setting
$\lambda_1=\lambda_{\mathrm{TV}}=1$ across the factorial grid of sample sizes
and nominal ranks. Figure~\ref{fig:robustness_sensitivity} summarizes explained
variance, loading sparsity, repeat stability, and convergence across the
evaluated $(N,K)$ settings.

Increasing $K$ progressively expands the reconstruction capacity of the
representation. At $N=20{,}000$, explained variance rises from $33.44\%$ at
$K=5$ to $40.55\%$ at $K=15$ and $43.98\%$ at $K=30$. The corresponding
trajectories are similar across all four sample sizes, showing consistent
reconstruction behaviour from $N=5{,}000$ to $N=20{,}000$.

The structural response is similarly systematic. As the nominal rank
increases, the representation distributes its support across a larger number
of components, while the sparsity trajectories remain smooth and consistent
across sample sizes. At $K=15$, mean sparsity ranges from $71.35\%$ for
$N=5{,}000$ to $62.46\%$ for $N=20{,}000$. This sample-size dependence is
consistent with the scaling of the objective: the reconstruction term
accumulates over $N$ observations, whereas the structural penalties contain no
explicit sample-size normalization. The resulting change in relative
regularization strength is quantified in
Appendix~\ref{app:objective_scaling}.

Component reproducibility strengthens with sample size. At $K=15$, all five
repeated fits converge for every evaluated $N$, and repeat stability increases
from $0.890$ at $N=5{,}000$ to $0.925$, $0.963$, and $0.996$ at
$N=10{,}000$, $15{,}000$, and $20{,}000$, respectively. The largest sample
therefore yields almost identical matched loading structures across repeated
fits.

\begin{figure*}[!ht]
    \centering

    \begin{subfigure}[t]{0.97\textwidth}
        \centering
        \includegraphics[
            width=\textwidth
        ]{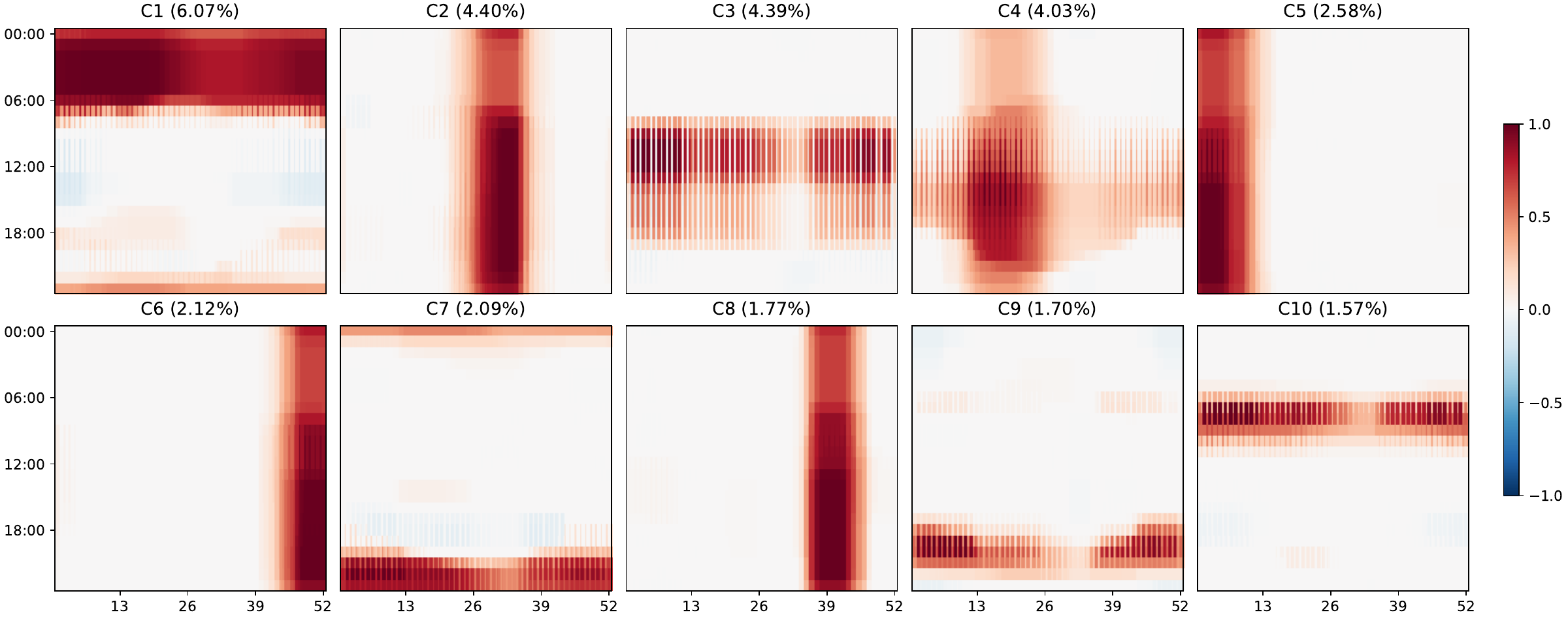}
        \caption{$K=10$.}
        \label{fig:rank_progression_k10}
    \end{subfigure}

    \vspace{0.8em}

    \begin{subfigure}[t]{0.97\textwidth}
        \centering
        \includegraphics[
            width=\textwidth
        ]{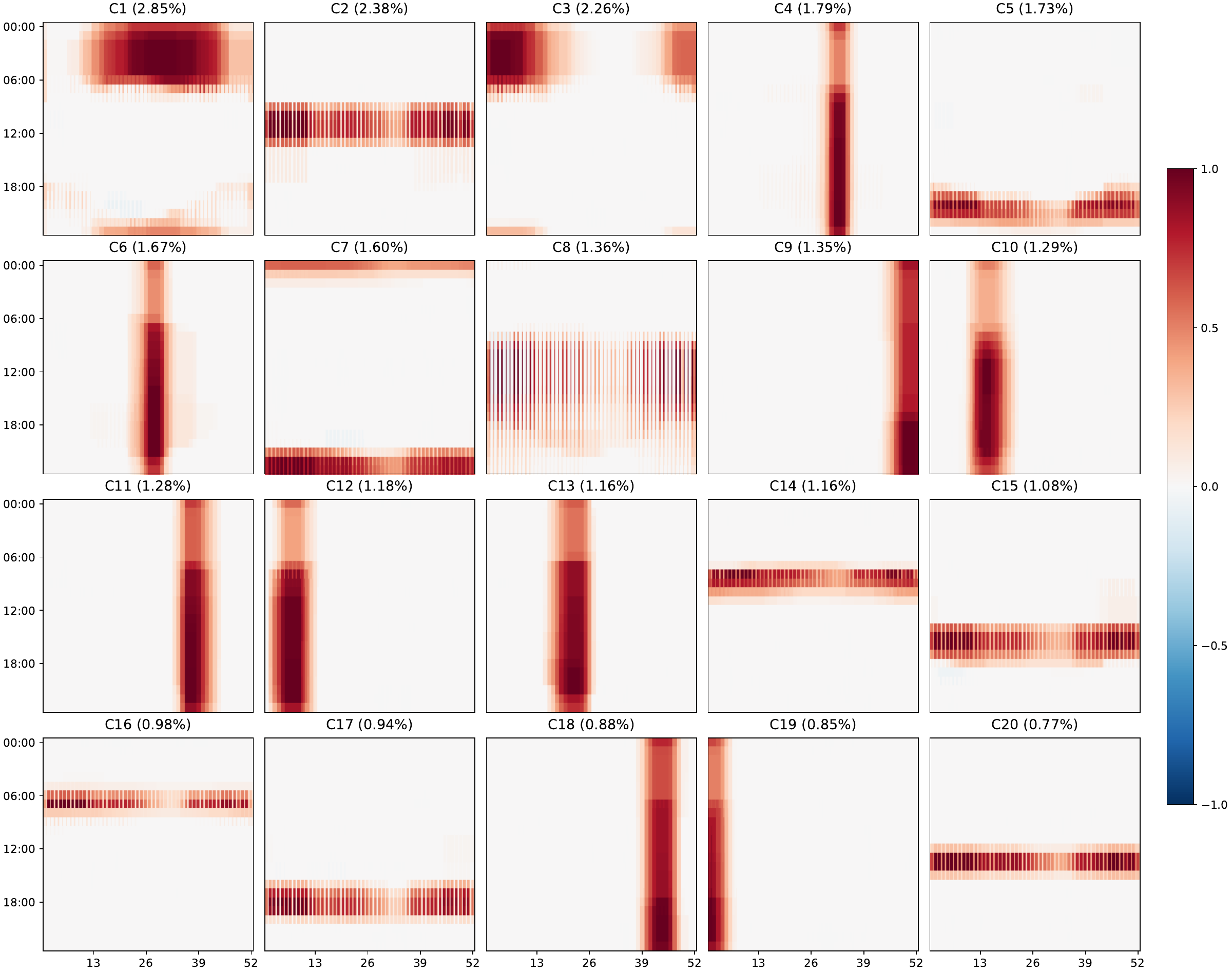}
        \caption{$K=20$.}
        \label{fig:rank_progression_k20}
    \end{subfigure}

    \caption{
    Illustrative Calendar-SPCA solutions for GoiEner at
    $N=20{,}000$ and $\lambda_1=\lambda_{\mathrm{TV}}=1$,
    showing the evolution of the learned representation with nominal rank.
    Components are ordered by decreasing conditional reconstruction
    contribution. The vertical axis represents time of day and the horizontal
    axis follows the 364 calendar days, grouped by ISO week. Each loading is
    independently normalized by its maximum absolute coefficient for
    visualization.
    }
    \label{fig:rank_progression}
\end{figure*}

The corresponding evolution of the loading structure is illustrated in
Figure~\ref{fig:rank_progression} using the $K=10$ and $K=20$ solutions from
the $N=20{,}000$ factorial setting. At $K=10$, the representation is organized
into relatively broad regions of the calendar domain. Increasing the rank
distributes the support across a larger number of progressively narrower and
more localized components. The visual progression therefore shows how nominal
rank controls the granularity of the learned representation.

Together, the quantitative results and the loading evolution place $K=15$
within a stable and reproducible operating region with substantial
reconstruction capacity and an intermediate level of structural granularity.
This rank is therefore adopted as the common nominal dimensionality for the
final GoiEner and LCL representations. With $K$ fixed, the next analysis
selects the regularization strength independently for each dataset.

\subsection{How is the operating point selected?}
\label{sec:results_lcurve}

\begin{figure*}[!ht]
    \centering
    \includegraphics[
        width=\textwidth
    ]{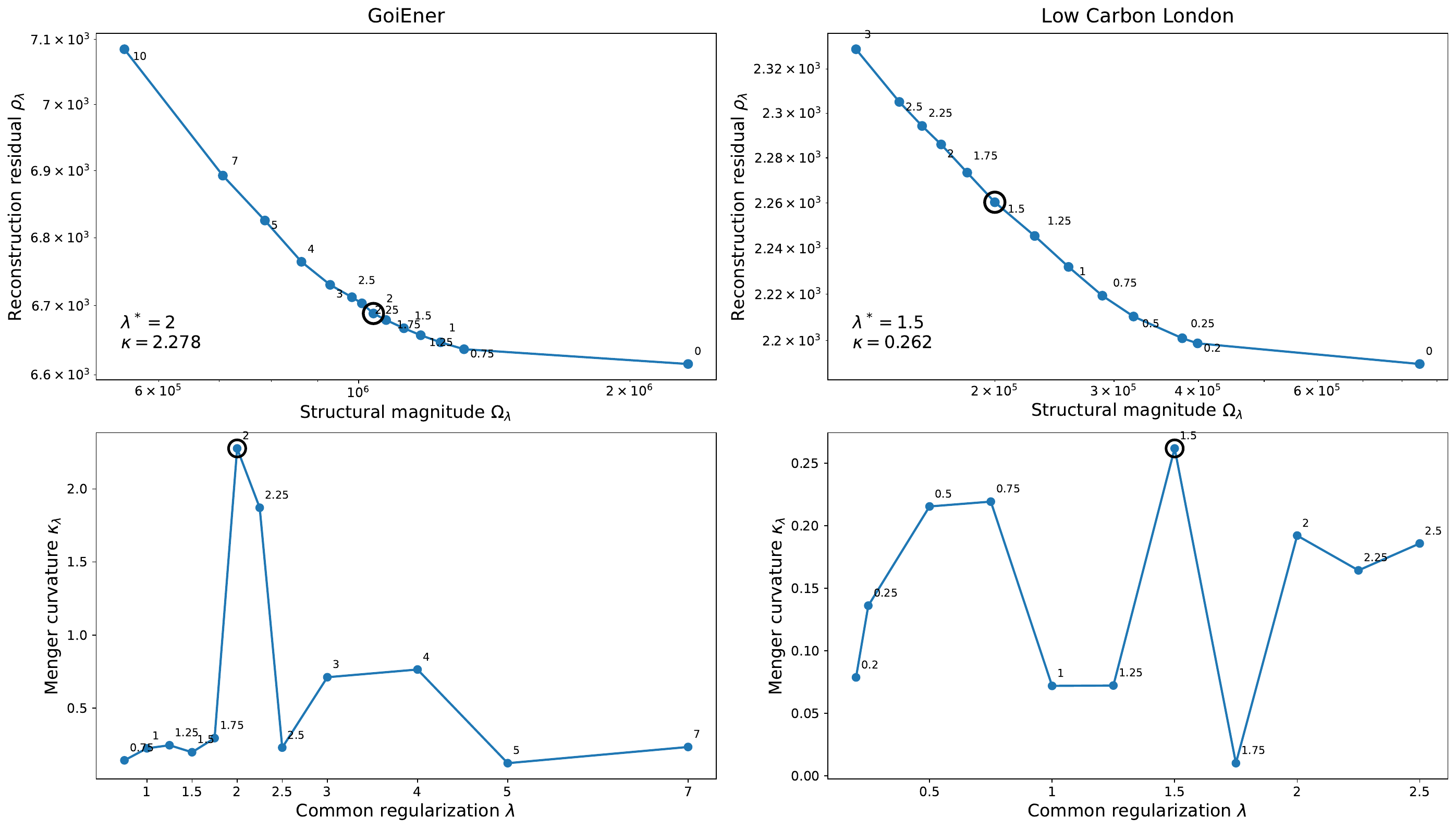}
    \caption{
    L-curve selection of the final rank-$15$ Calendar-SPCA operating points.
    (a,b) Reconstruction--structure trade-off for GoiEner and Low Carbon
    London along $\lambda_1=\lambda_{\mathrm{TV}}=\lambda$.
    (c,d) Discrete curvature of the corresponding eligible paths.
    Maximum curvature selects $\lambda^*=2.0$ for GoiEner and
    $\lambda^*=1.5$ for Low Carbon London.
    }
    \label{fig:lcurve_selection}
\end{figure*}

With $K=15$ fixed, the regularization strength was selected independently for
GoiEner and LCL along the diagonal path
$\lambda_1=\lambda_{\mathrm{TV}}=\lambda$.
Figure~\ref{fig:lcurve_selection} shows the resulting L-curves and their
discrete curvature.

Increasing $\lambda$ progressively strengthens the structural regularization,
reducing the structural magnitude of the loading representation while
increasing the reconstruction residual. This produces a clear regularization
path in both datasets, whose curvature identifies the region of strongest
transition in the reconstruction--structure trade-off.

The maximum curvature occurs at

\begin{equation}
\begin{aligned}
    \text{GoiEner:}\qquad
    &\lambda^*=2.0,\\
    \text{Low Carbon London:}\qquad
    &\lambda^*=1.5.
\end{aligned}
\end{equation}

Both selected points retain the complete nominal rank
$K_{\mathrm{eff}}=15$ and define the final Calendar-SPCA configurations for
the two datasets.

The L-curve criterion provides a reproducible operating-point rule for the
comparative evaluation. In application-specific settings, the regularization
path can also be selected according to the desired balance among
reconstruction, sparsity, calendar coherence, and temporal localization.
With both $K$ and the dataset-specific regularization strengths established,
the next analysis examines the structure learned by the final representations.

\subsection{What does Calendar-SPCA learn?}
\label{sec:results_interpretable_components}

\begin{figure*}[!p]
    \centering

    \begin{subfigure}[t]{0.97\textwidth}
        \centering
        \includegraphics[
            width=\textwidth
        ]{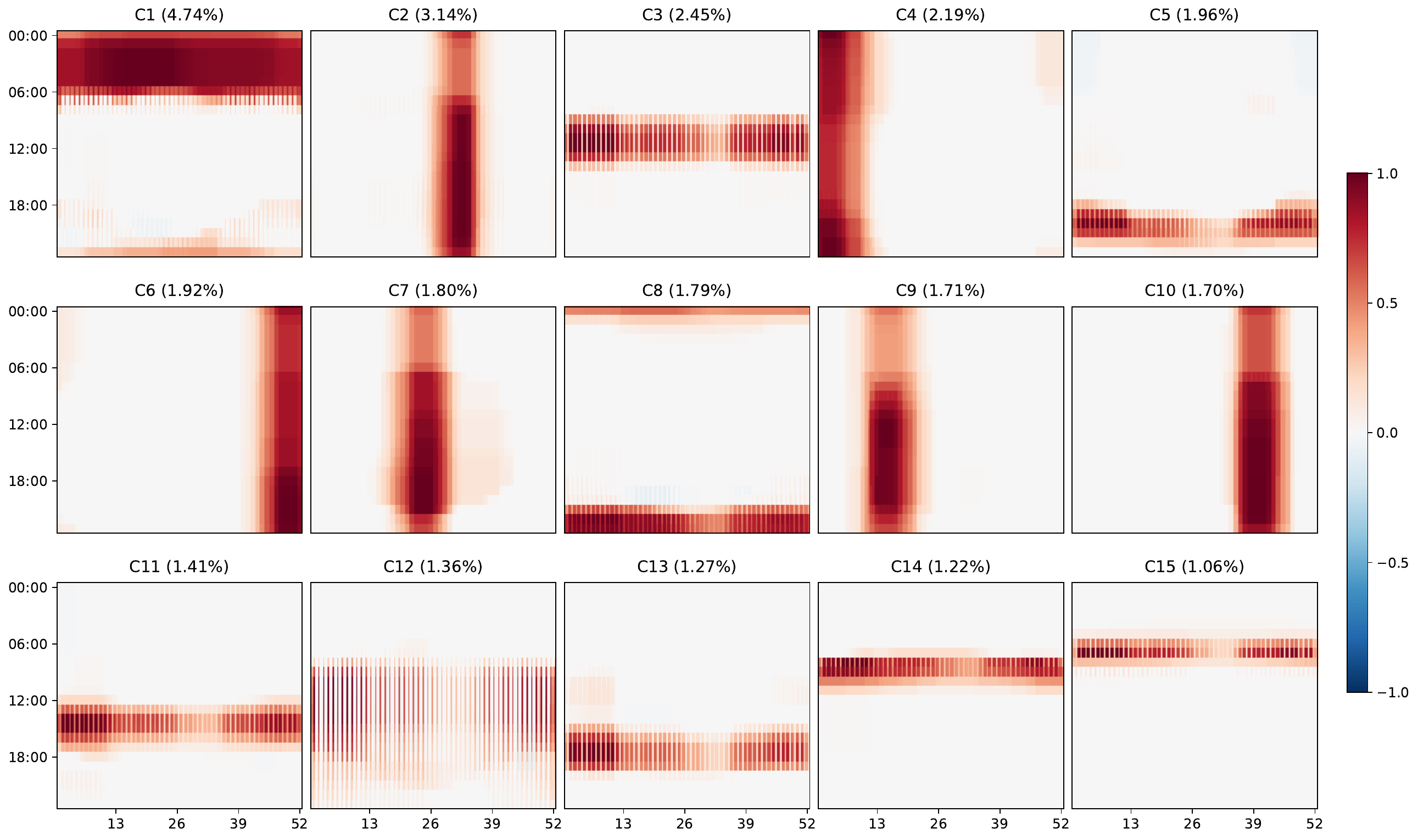}
        \caption{GoiEner (\(K=15\), \(\lambda^*=2.0\)).}
        \label{fig:components_goiener}
    \end{subfigure}

    \vspace{0.8em}

    \begin{subfigure}[t]{0.97\textwidth}
        \centering
        \includegraphics[
            width=\textwidth
        ]{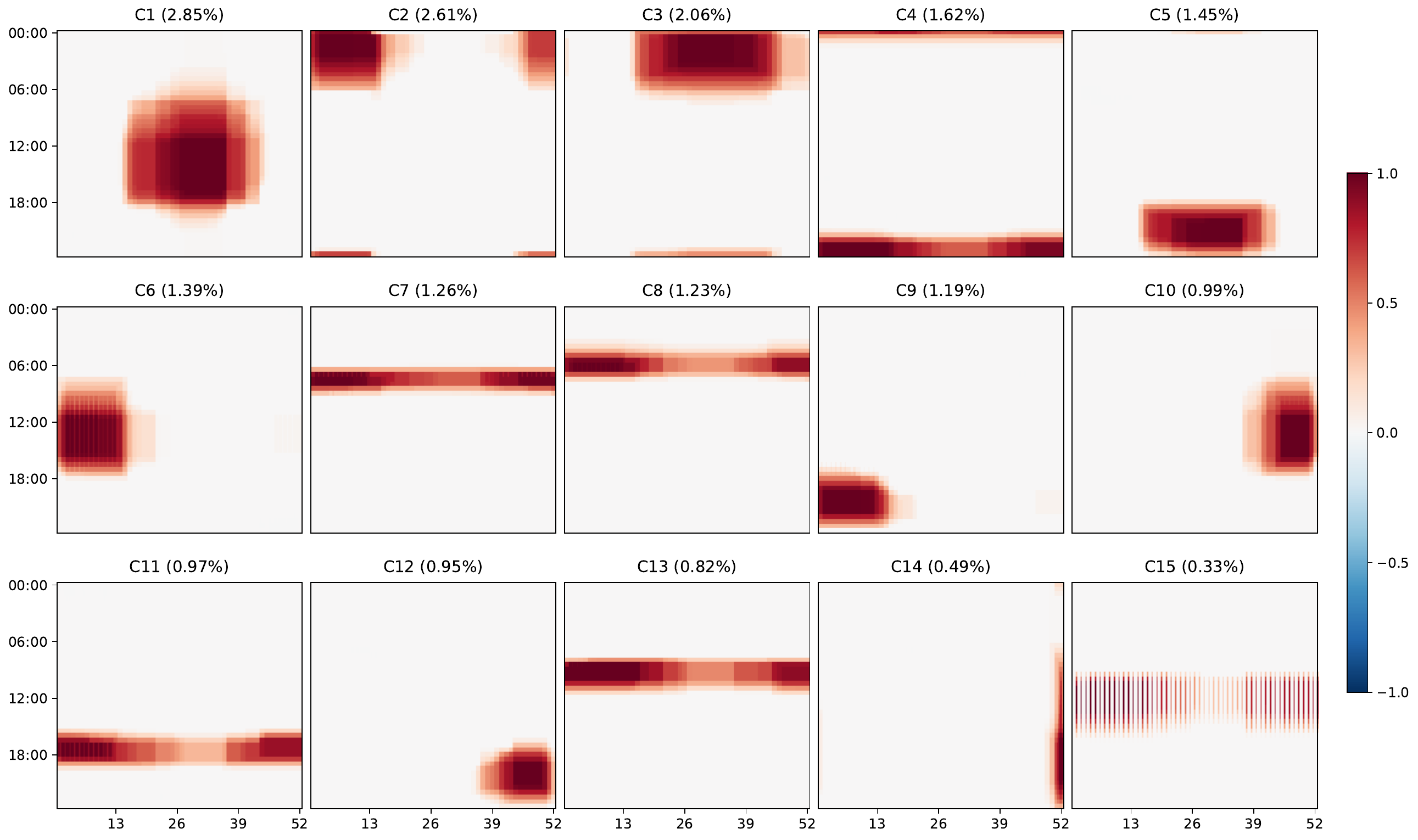}
        \caption{Low Carbon London (\(K=15\), \(\lambda^*=1.5\)).}
        \label{fig:components_lcl}
    \end{subfigure}

    \caption{
        Complete rank-$15$ Calendar-SPCA representations for
        (a) GoiEner and
        (b) Low Carbon London.
        Components are ordered by decreasing conditional reconstruction
        contribution, reported in parentheses.
        The vertical axis represents time of day and the horizontal axis
        follows the 364 calendar days, grouped by ISO week.
        Each loading is independently normalized by its maximum absolute
        coefficient for visualization.
    }
    \label{fig:selected_components}
\end{figure*}

Using the operating points selected above, the final Calendar-SPCA solutions
reveal how the learned factors occupy the original calendar domain.
Figure~\ref{fig:selected_components} shows the complete rank-$15$ loading
systems obtained for GoiEner and Low Carbon London, with components ordered by
decreasing conditional reconstruction contribution. Because the loadings are
displayed directly in their temporal coordinates, daily, weekly, seasonal,
and jointly localized structures can be identified from the component maps
themselves.

The GoiEner representation is strongly organized along recognizable calendar
directions. Several components describe distinct portions of the
\emph{daily cycle}. C1 captures a night-time pattern extending across all days
of the week, while other daily factors are more specifically associated with
weekdays: C15 and C14 emphasize morning hours, C3 and C11 capture midday and
afternoon activity, and C13, C5, and C8 represent progressively later parts
of the day. A second group separates the \emph{annual cycle} into broad
seasonal intervals, with C4, C9, C7, C2, C10, and C6 concentrating on
successive portions of the year. The \emph{weekly cycle} is represented
particularly clearly by C12, whose recurrent pattern isolates weekend
behaviour throughout the annual cycle. Together, these components form a
decomposition in which daily, weekly, and seasonal organization can be
identified directly from the loading maps.

The Low Carbon London representation follows the same calendar principle,
with many factors localized jointly across several temporal coordinates. C1,
for example, isolates daytime hours during the middle part of the year, while
C6 and C10 capture related daytime patterns concentrated toward earlier and
later seasonal intervals. Evening and night-time behaviour is similarly
separated into temporally localized factors, including C5, C9, and C12.
Other components, such as C7, C8, C11, and C13, extend over broader portions
of the annual cycle while remaining concentrated within particular within-day
intervals. The \emph{weekly cycle} is particularly evident in C15, whose
recurrent pattern isolates weekend behaviour within a restricted daytime
window. Together, these components span both axis-dominated structures and
more specific combinations of daily, weekly, and annual coordinates.

The two datasets therefore illustrate complementary organizations within the
same calendar geometry. GoiEner contains many factors aligned predominantly
with individual calendar axes, while LCL contains more factors localized
jointly across several temporal coordinates. In both cases, the location and
extent of the active regions are learned directly from the data.

The cyclic topology is also visible in the learned components and is
particularly clear in LCL component C2. Its loading extends simultaneously
across the week-52/week-1 boundary of the annual axis and across the midnight
boundary of the within-day axis. Although these regions appear separated at
the edges of the displayed heatmap, they form a contiguous structure on the
cyclic calendar graph.

These final representations provide the structural reference for the
cross-method comparison in the next subsection.

\subsection{How does Calendar-SPCA compare with existing methods?}
\label{sec:results_representation_quality}

Having established the structure of the final Calendar-SPCA representations,
we compare them with PCA, SPCA, and SPCA-TV on the complete GoiEner and Low
Carbon London datasets. Table~\ref{tab:method_comparison} summarizes
reconstruction retention, loading sparsity, and calendar coherence for the
three structured representations, using rank-matched PCA as the reconstruction
reference.

\begin{table}[!ht]
\centering
\caption{
Quantitative comparison of the rank-$15$ representations.
Explained-variance retention is measured relative to rank-matched PCA.
Sparsity is the mean proportion of numerically zero loading coefficients, and RTV is the 
mean relative total variation over the common cyclic calendar graph;
lower RTV indicates greater local coherence.
}
\label{tab:method_comparison}

\begin{tabular}{llccc}
\toprule
Dataset & Method & EV retention (\%) & Sparsity (\%) & RTV \\
\midrule
& SPCA          & 98.44 & 47.12 & 0.994 \\
GoiEner
& SPCA-TV       & 81.98 & 42.10 & 0.792 \\
& Calendar-SPCA & 96.92 & 61.95 & 0.615 \\
\midrule
& SPCA          & 94.63 & 64.24 & 1.052 \\
LCL
& SPCA-TV       & 93.22 & 19.02 & 0.509 \\
& Calendar-SPCA & 82.90 & 81.50 & 0.450 \\
\bottomrule
\end{tabular}

\end{table}

Calendar-SPCA achieves the highest mean sparsity and the lowest RTV of the
three structured representations on both datasets. At the selected operating
points, it retains $96.92\%$ of rank-matched PCA explained variance on GoiEner
and $82.90\%$ on Low Carbon London, while concentrating the loadings into
substantially more selective and locally coherent calendar regions.

\begin{figure*}[!ht]
\centering

\begin{subfigure}[t]{0.97\textwidth}
    \centering
    \includegraphics[
        width=\textwidth
    ]{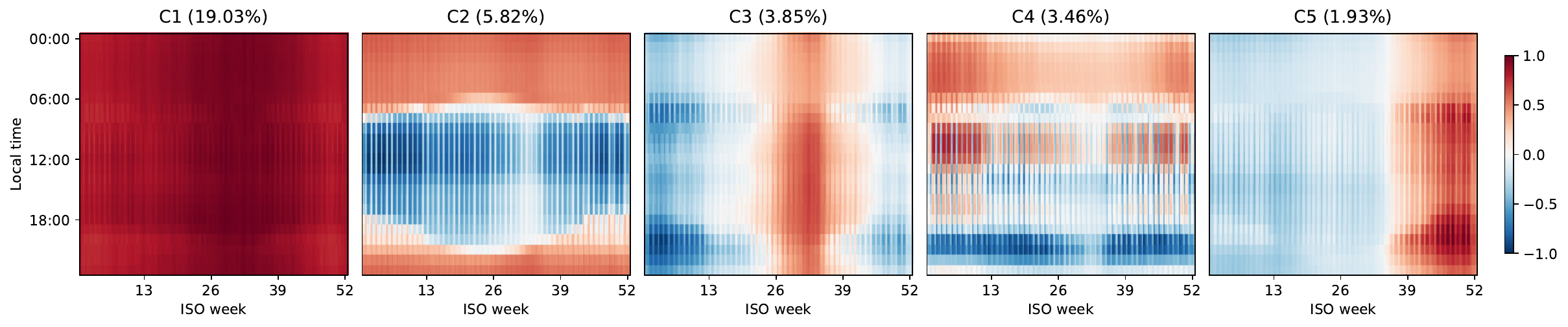}
    \caption{PCA.}
    \label{fig:baseline_first5_pca}
\end{subfigure}

\vspace{0.8em}

\begin{subfigure}[t]{0.97\textwidth}
    \centering
    \includegraphics[
        width=\textwidth
    ]{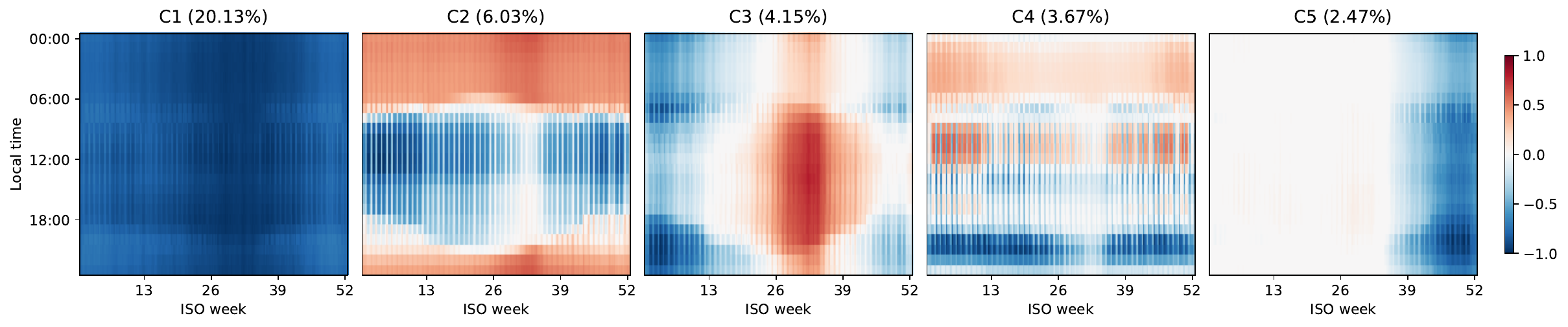}
    \caption{Sparse PCA.}
    \label{fig:baseline_first5_spca}
\end{subfigure}

\vspace{0.8em}

\begin{subfigure}[t]{0.97\textwidth}
    \centering
    \includegraphics[
        width=\textwidth
    ]{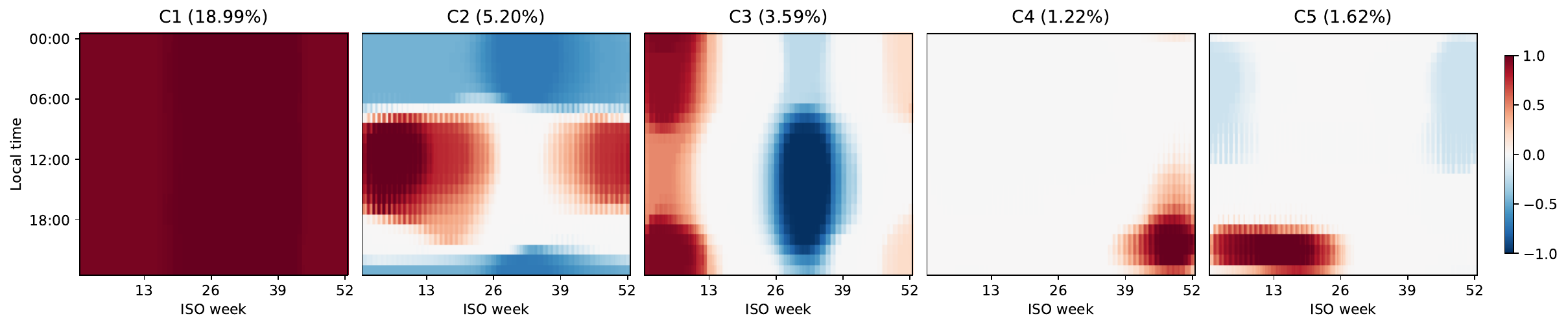}
    \caption{SPCA-TV.}
    \label{fig:baseline_first5_spcatv}
\end{subfigure}

\caption{
First five components of the rank-$15$ PCA, SPCA, and SPCA-TV
representations fitted to the complete GoiEner dataset.
Components are shown in the native order returned by each method.
The vertical axis represents time of day and the horizontal axis follows
the 364 calendar days, grouped by ISO week. Each loading is independently
normalized by its maximum absolute coefficient for visualization.
}
\label{fig:baseline_first5}

\end{figure*}

The aggregate metrics quantify the reconstruction--structure trade-off, while
the component maps reveal how the latent coordinates are organized over the
calendar domain.
For illustrative purposes, Figure~\ref{fig:baseline_first5}
shows the first five components of the rank-$15$ PCA, SPCA, and SPCA-TV
representations fitted to the complete GoiEner dataset. Components are shown
in the native order returned by each method, without additional selection or
reordering.

PCA and SPCA produce closely related component systems. Their components
combine positive and negative coefficients over broad temporal regions with
weakly defined boundaries. SPCA largely preserves the PCA organization while
setting selected parts of the loadings to zero. Consequently, individual
components remain difficult to associate with a specific time-of-day, weekly,
or seasonal factor, consistent with the well-known interpretability limitations
of PCA-based representations. In particular, the leading component is
essentially a global intensity mode spanning the entire calendar and, because
it does not isolate any specific region, is largely uninformative as a factor
of temporal structure.

SPCA-TV produces a more localized representation. Its first component is again
a global calendar mode. Components C2 and C3 contain distinct positive and
negative regions within the same loading, making their temporal interpretation
less direct. In contrast, C4 and all subsequent components exhibit clear
localization along the annual and within-day coordinates. However, several of
these components are reduced to small patches confined to narrow portions of
the calendar, yielding a fragmented decomposition that does not provide
complete coverage of the temporal domain.

Taken together, the numerical and visual comparisons show that Calendar-SPCA
organizes sparsity and local coherence into a complementary set of temporal
factors spanning the calendar domain. This organization preserves substantial
reconstruction capacity while providing the directly identifiable daily,
weekly, seasonal, and jointly localized structures described above. The full
numerical comparison is reported in Appendix~\ref{app:baseline_summary}.


\section{Discussion}
\label{sec:discussion}
The central contribution of Calendar-SPCA is the direct alignment between the
geometry used during learning and the temporal coordinates used to interpret
the resulting representation. The calendar graph specifies which positions are
neighbours, while the optimization determines which regions become active,
their extent, and how several temporal coordinates combine within each factor.
Calendar structure therefore acts as a geometric prior that guides the
representation while preserving data-driven flexibility in the organization
of the latent factors.

This alignment also gives the two structural regularizers a complementary role
within the latent system. Loading sparsity controls temporal selectivity,
while graph total variation promotes local coherence over the calendar
topology. Their joint estimation coordinates these properties across a common
set of factors, producing latent dimensions that retain substantial population
information while remaining expressed in meaningful temporal coordinates.
This combination is particularly useful when the learned representation is
intended to support subsequent analysis as a set of interpretable temporal
variables.

The score matrix $U$ provides the complementary representation at observation
level. Each row defines a compact $K$-dimensional embedding of an annual
profile in the learned calendar-factor space, with each coordinate measuring
its expression along one latent temporal factor. These embeddings can support
consumer clustering and segmentation, comparison of consumption populations,
visualization of large smart-meter collections, and association with
household, technological, socioeconomic, or other available metadata. They
can also serve as inputs to downstream classification, regression,
forecasting, or anomaly-detection pipelines. In this representation, the
loading matrix $V$ describes the temporal meaning of the latent dimensions,
while the score matrix $U$ describes how strongly each observation expresses
them, providing complementary temporal and observation-level views of the
same latent space.

The calendar construction also extends naturally beyond the three axes used
in the present experiments. A feature domain containing $p$ known periodic
coordinates can be represented through a Cartesian product of cycle graphs,

\begin{equation}
    G
    =
    C_{m_1}
    \mathbin{\square}
    C_{m_2}
    \mathbin{\square}
    \cdots
    \mathbin{\square}
    C_{m_p},
\end{equation}

allowing four-, five-, or higher-dimensional periodic representations whenever
the application provides the corresponding coordinates. Multi-periodic
structure appears in many domains beyond electricity consumption. Daily and
weekly patterns have also been reported in telecommunications traffic
\cite{vargasanamuro2024lte} and industrial activity
\cite{lee2022manufacturing}; traffic profiles vary across time of day, day of
week, and season \cite{crawford2017traffic}; building occupancy follows daily,
weekly, and calendar schedules \cite{alfalah2023occupancy}; and ecological
observations can combine diel, tidal, and seasonal cycles
\cite{fernandezbetelu2019dolphins}. These settings share the same underlying
idea: known periodic coordinates define a structured feature domain that can
be incorporated directly into representation learning. Calendar-SPCA provides
a general mechanism for expressing that geometry through sparse,
graph-coherent latent factors.


\section{Conclusions}
\label{sec:conclusions}

This work introduced Calendar-SPCA, a structured sparse representation-learning
method for long-term multi-periodic electricity-consumption profiles. The
method represents the feature domain through the Cartesian product of cyclic
calendar axes and combines sparse loading estimation with graph total
variation within a joint low-rank factorization. The resulting components are
learned directly on daily, weekly, and annual coordinates and can be
interpreted as coherent temporal patterns on the original calendar domain.

Experiments on the GoiEner and Low Carbon London datasets demonstrated
controllable sparsity and calendar coherence, systematic behaviour across
sample sizes and latent dimensionalities, high repeat stability within the
identified operating region, reproducible operating-point selection, and
substantial retention of the variance captured by rank-matched PCA. The recovered components reveal readable daily, weekly, seasonal, and
jointly localized patterns across datasets with different population sizes
and temporal resolutions. More broadly, Calendar-SPCA illustrates how known
feature geometry can be incorporated directly into representation learning to
obtain latent factors whose structure is aligned with the coordinates used to
understand the data.


\section*{Data availability}

The GoiEner Smart Meter Dataset (Version 7.1) used in this study is publicly
available through Zenodo \cite{quesada2026goiener}. The Low Carbon London
smart-meter dataset is publicly available through the London Datastore
\cite{ukpowernetworks_lcl}.

\section*{Code availability}

The implementation of Calendar-SPCA and the code for the principal experiments reported in this study are available at \url{https://github.com/quesadagranja/calendar-spca}.

\section*{CRediT authorship contribution statement}

\noindent
\textbf{Carlos Quesada-Granja:}
Conceptualization, Methodology, Software, Validation, Formal analysis,
Investigation, Data curation, Visualization, Writing -- original draft,
Writing -- review \& editing, Supervision.

\vspace{0.4em}
\noindent
\textbf{Tony Castillo-Calzadilla:}
Writing -- original draft, Writing -- review \& editing.

\vspace{0.4em}
\noindent
\textbf{Carlos Rizo-Maestre:}
Writing -- original draft, Writing -- review \& editing.

\section*{Funding}

This research received no external funding.

\section*{Declaration of competing interest}

The authors declare no competing interests.


\bibliographystyle{unsrt}
\bibliography{references}

@article{alfalah2023occupancy,
  author  = {Alfalah, Bashar and Shahrestani, Mehdi and Shao, Li},
  title   = {Identifying occupancy patterns and profiles in higher education
             institution buildings with high occupancy density -- A case study},
  journal = {Intelligent Buildings International},
  volume  = {15},
  number  = {2},
  pages   = {45--61},
  year    = {2023},
  doi     = {10.1080/17508975.2022.2137451}
}

@article{banin2025predicting,
  title={Predicting energy demand with tensor factor models},
  author={Banin, Mattia and Barigozzi, Matteo and Trapin, Luca},
  journal={arXiv preprint arXiv:2502.06213},
  year={2025}
}

@article{condat2013primal,
  title={A primal--dual splitting method for convex optimization involving Lipschitzian, proximable and linear composite terms},
  author={Condat, Laurent},
  journal={Journal of optimization theory and applications},
  volume={158},
  number={2},
  pages={460--479},
  year={2013},
  publisher={Springer}
}

@article{crawford2017traffic,
  author  = {Crawford, Fiona and Watling, David P. and Connors, Richard D.},
  title   = {A statistical method for estimating predictable differences
             between daily traffic flow profiles},
  journal = {Transportation Research Part B: Methodological},
  volume  = {95},
  pages   = {196--213},
  year    = {2017},
  doi     = {10.1016/j.trb.2016.11.004}
}

@article{depierrefeu2018spcatv,
  title={Structured sparse principal components analysis with the TV-elastic net penalty},
  author={De Pierrefeu, Amicie and L{\"o}fstedt, Tommy and Hadj-Selem, Fouad and Dubois, Mathieu and Jardri, Renaud and Fovet, Thomas and Ciuciu, Philippe and Frouin, Vincent and Duchesnay, Edouard},
  journal={IEEE transactions on medical imaging},
  volume={37},
  number={2},
  pages={396--407},
  year={2017},
  publisher={IEEE}
}

@article{fernandezbetelu2019dolphins,
  author  = {Fernandez-Betelu, Oihane and Graham, Isla M. and
             Cornulier, Thomas and Thompson, Paul M.},
  title   = {Fine scale spatial variability in the influence of environmental
             cycles on the occurrence of dolphins at coastal sites},
  journal = {Scientific Reports},
  volume  = {9},
  pages   = {2548},
  year    = {2019},
  doi     = {10.1038/s41598-019-38900-4}
}

@article{guo2010principal,
  title={Principal component analysis with sparse fused loadings},
  author={Guo, Jian and James, Gareth and Levina, Elizaveta and Michailidis, George and Zhu, Ji},
  journal={Journal of Computational and Graphical Statistics},
  volume={19},
  number={4},
  pages={930--946},
  year={2010},
  publisher={Taylor \& Francis}
}

@article{hansen1992analysis,
  title={Analysis of discrete ill-posed problems by means of the L-curve},
  author={Hansen, Per Christian},
  journal={SIAM review},
  volume={34},
  number={4},
  pages={561--580},
  year={1992},
  publisher={SIAM}
}

@article{hansen1993use,
  title={The use of the L-curve in the regularization of discrete ill-posed problems},
  author={Hansen, Per Christian and O’Leary, Dianne Prost},
  journal={SIAM journal on scientific computing},
  volume={14},
  number={6},
  pages={1487--1503},
  year={1993},
  publisher={SIAM}
}

@inproceedings{jenatton2010structured,
  title={Structured sparse principal component analysis},
  author={Jenatton, Rodolphe and Obozinski, Guillaume and Bach, Francis},
  booktitle={Proceedings of the Thirteenth International Conference on Artificial Intelligence and Statistics},
  pages={366--373},
  year={2010},
  organization={JMLR Workshop and Conference Proceedings}
}

@article{jolliffe2003modified,
  title={A modified principal component technique based on the LASSO},
  author={Jolliffe, Ian T and Trendafilov, Nickolay T and Uddin, Mudassir},
  journal={Journal of computational and Graphical Statistics},
  volume={12},
  number={3},
  pages={531--547},
  year={2003},
  publisher={Taylor \& Francis}
}

@article{jolliffe2016principal,
  title={Principal component analysis: a review and recent developments},
  author={Jolliffe, Ian T and Cadima, Jorge},
  journal={Philosophical transactions. Series A, Mathematical, physical, and engineering sciences},
  volume={374},
  number={2065},
  pages={20150202},
  year={2016}
}

@article{kolda2009tensor,
  title={Tensor decompositions and applications},
  author={Kolda, Tamara G and Bader, Brett W},
  journal={SIAM review},
  volume={51},
  number={3},
  pages={455--500},
  year={2009},
  publisher={SIAM}
}

@article{lee2022manufacturing,
  author  = {Lee, Eunjung and Baek, Keon and Kim, Jinho},
  title   = {Datasets on South Korean manufacturing factories' electricity
             consumption and demand response participation},
  journal = {Scientific Data},
  volume  = {9},
  pages   = {227},
  year    = {2022},
  doi     = {10.1038/s41597-022-01357-8}
}

@article{li2017incorporating,
  title={Incorporating biological information in sparse principal component analysis with application to genomic data},
  author={Li, Ziyi and Safo, Sandra E and Long, Qi},
  journal={BMC bioinformatics},
  volume={18},
  number={1},
  pages={332},
  year={2017},
  publisher={Springer}
}

@article{quesada2024electricity,
  title={An electricity smart meter dataset of Spanish households: insights into consumption patterns},
  author={Quesada, Carlos and Astigarraga, Leire and Merveille, Chris and Borges, Cruz E},
  journal={Scientific Data},
  volume={11},
  number={1},
  pages={59},
  year={2024},
  publisher={Nature Publishing Group UK London}
}

@article{quesada2025data,
  title={A data-driven methodology for deriving electricity consumption typologies from smart meters},
  author={Quesada, Carlos and Montero-Manso, Pablo and Pflugradt, Noah and Astigarraga, Leire and Merveille, Chris and Casado-Mansilla, Diego and Borges, Cruz E},
  journal={Energy Reports},
  volume={14},
  pages={2420--2434},
  year={2025},
  publisher={Elsevier}
}

@misc{quesada2026goiener,
  author       = {Quesada-Granja, Carlos},
  title        = {{GoiEner Smart Meter Dataset} (Version 7.1)},
  year         = {2026},
  howpublished = {Zenodo},
  note         = {\url{https://doi.org/10.5281/zenodo.22313409}},
  doi          = {10.5281/zenodo.22313409}
}

@article{shen2008sparse,
  title={Sparse principal component analysis via regularized low rank matrix approximation},
  author={Shen, Haipeng and Huang, Jianhua Z},
  journal={Journal of multivariate analysis},
  volume={99},
  number={6},
  pages={1015--1034},
  year={2008},
  publisher={Elsevier}
}

@article{tibshirani2005sparsity,
  title={Sparsity and smoothness via the fused lasso},
  author={Tibshirani, Robert and Saunders, Michael and Rosset, Saharon and Zhu, Ji and Knight, Keith},
  journal={Journal of the Royal Statistical Society Series B: Statistical Methodology},
  volume={67},
  number={1},
  pages={91--108},
  year={2005},
  publisher={Oxford University Press}
}

@book{tibshirani2011solution,
  title={The solution path of the generalized lasso},
  author={Tibshirani, Ryan J},
  year={2011},
  publisher={Stanford University}
}

@book{tukey1977exploratory,
  title={Exploratory data analysis},
  author={Tukey, John Wilder and others},
  volume={2},
  year={1977},
  publisher={Addison-wesley Reading, MA}
}

@misc{ukpowernetworks_lcl,
  author       = {{UK Power Networks}},
  title        = {{SmartMeter Energy Consumption Data in London Households}},
  howpublished = {London Datastore},
  note         = {\url{https://data.london.gov.uk/dataset/smartmeter-energy-consumption-data-in-london-households-vqm0d}, accessed 5 September 2026}
}

@article{vargasanamuro2024lte,
  author  = {Vargas Anamuro, Cesar and Blanc, Alberto and Lagrange, Xavier},
  title   = {Statistical analysis and characterization of signaling and user
             traffic of a commercial multi-band {LTE} system},
  journal = {Telecommunication Systems},
  volume  = {87},
  number  = {2},
  pages   = {437--453},
  year    = {2024},
  doi     = {10.1007/s11235-024-01196-5}
}

@article{vu2013splitting,
  title={A splitting algorithm for dual monotone inclusions involving cocoercive operators},
  author={V{\~u}, B{\`{\u{a}}}ng C{\^o}ng},
  journal={Advances in Computational Mathematics},
  volume={38},
  number={3},
  pages={667--681},
  year={2013},
  publisher={Springer}
}

@article{witten2009penalized,
  title={A penalized matrix decomposition, with applications to sparse principal components and canonical correlation analysis},
  author={Witten, Daniela M and Tibshirani, Robert and Hastie, Trevor},
  journal={Biostatistics},
  volume={10},
  number={3},
  pages={515--534},
  year={2009},
  publisher={Oxford University Press}
}

@article{zou2006sparse,
  title={Sparse principal component analysis},
  author={Zou, Hui and Hastie, Trevor and Tibshirani, Robert},
  journal={Journal of computational and graphical statistics},
  volume={15},
  number={2},
  pages={265--286},
  year={2006},
  publisher={Taylor \& Francis}
}

\clearpage
\appendix


\section{Optimization details}
\label{app:optimization}

This appendix completes the numerical specification of the Calendar-SPCA
fitting procedure described in Section~\ref{sec:optimization}. It provides the
primal--dual loading updates, step-size selection, convergence criteria,
initialization and component-recovery strategy, and the numerical settings used
throughout the experiments.

\subsection{Primal--dual loading update}
\label{app:loading_update}

For fixed $U$, the loading matrix is obtained by solving the convex problem

\begin{equation}
\min_V
\left\{
\frac{1}{2}
\|X_c-UV^\top\|_F^2
+
\lambda_1\|V\|_{1,1}
+
\lambda_{\mathrm{TV}}\|D_GV\|_{1,1}
\right\}.
\label{eq:appendix_loading_subproblem}
\end{equation}

The smooth reconstruction term

\[
f(V)
=
\frac{1}{2}
\|X_c-UV^\top\|_F^2
\]

has gradient

\begin{equation}
    \nabla f(V)
    =
    V(U^\top U)-X_c^\top U.
    \label{eq:appendix_loading_gradient}
\end{equation}

Calendar-SPCA solves
Eq.~\eqref{eq:appendix_loading_subproblem} with a Condat--V\~u
primal--dual splitting scheme \cite{condat2013primal,vu2013splitting}. Introducing a dual
variable $Y$ associated with the calendar incidence operator $D_G$, one
iteration is

\begin{equation}
V^{(\ell+1)}
=
\operatorname{soft}
\left(
V^{(\ell)}
-
\tau
\left[
\nabla f\!\left(V^{(\ell)}\right)
+
D_G^\top Y^{(\ell)}
\right],
\tau\lambda_1
\right),
\label{eq:primal_update}
\end{equation}

followed by the extrapolation

\begin{equation}
    \overline V^{(\ell+1)}
    =
    2V^{(\ell+1)}-V^{(\ell)},
    \label{eq:primal_extrapolation}
\end{equation}

and the dual update

\begin{equation}
Y^{(\ell+1)}
=
\operatorname{clip}
\left(
Y^{(\ell)}
+
\sigma D_G\overline V^{(\ell+1)},
-\lambda_{\mathrm{TV}},
\lambda_{\mathrm{TV}}
\right).
\label{eq:dual_update}
\end{equation}

Here, $\operatorname{soft}(z,\theta)$ denotes entrywise soft-thresholding,

\begin{equation}
    \operatorname{soft}(z,\theta)
    =
    \operatorname{sign}(z)
    \max\{|z|-\theta,0\},
\end{equation}

and $\operatorname{clip}$ denotes entrywise projection onto the interval
$[-\lambda_{\mathrm{TV}},\lambda_{\mathrm{TV}}]$.

The primal update therefore applies the sparsity penalty directly to the
loading coefficients, while the dual update controls differences across the
calendar graph through $D_G$. The two regularizers are handled within the same
convex loading solve, preserving the sparse and graph-coherent structure of
the Calendar-SPCA objective.

The primal and dual step sizes $\tau$ and $\sigma$ are chosen from the
spectral properties of $U$ and $D_G$, as described next.

\subsection{Step-size selection}
\label{app:step_sizes}

The primal--dual step sizes depend on the Lipschitz constant of the smooth
loading objective and on the spectral norm of the calendar incidence
operator.

For fixed $U$, the gradient in
Eq.~\eqref{eq:appendix_loading_gradient} has Lipschitz constant

\begin{equation}
    L_U
    =
    \lambda_{\max}(U^\top U).
    \label{eq:loading_lipschitz}
\end{equation}

The calendar incidence matrix satisfies

\begin{equation}
    D_G^\top D_G
    =
    L_G,
\end{equation}

where $L_G$ is the graph Laplacian. For the Cartesian-product calendar graph

\[
G
=
C_h
\mathbin{\square}
C_d
\mathbin{\square}
C_w,
\]

the Laplacian eigenvalues are obtained by summing the eigenvalues of the three
cycle graphs. Consequently,

\begin{equation}
    \|D_G\|_2^2
    =
    \ell(h)+\ell(d)+\ell(w),
    \label{eq:calendar_incidence_norm}
\end{equation}

where the largest Laplacian eigenvalue of a cycle $C_m$ is

\begin{equation}
\ell(m)
=
\begin{cases}
4,
& m \text{ even},\\[2mm]
2+2\cos\left(\dfrac{\pi}{m}\right),
& m \text{ odd}.
\end{cases}
\label{eq:cycle_laplacian_max}
\end{equation}

Both calendar domains used in this study,
$(h,d,w)=(24,7,52)$ for GoiEner and $(48,7,52)$ for Low Carbon London,
therefore give

\begin{equation}
    \|D_G\|_2^2
    =
    10+2\cos\left(\frac{\pi}{7}\right)
    \approx
    11.80194.
    \label{eq:calendar_incidence_norm_value}
\end{equation}

Using $L_U$ and $\|D_G\|_2^2$, the dual and primal step sizes are chosen as

\begin{equation}
    \sigma
    =
    \frac{\gamma}{\sqrt{\|D_G\|_2^2}},
    \qquad
    \tau
    =
    \frac{\eta}
    {L_U/2+\sigma\|D_G\|_2^2},
    \label{eq:primal_dual_steps}
\end{equation}

with $\gamma>0$ controlling the dual step and $0<\eta<1$ providing a
numerical safety margin.

These choices satisfy the Condat--V\~u condition

\begin{equation}
    \frac{1}{\tau}
    -
    \sigma\|D_G\|_2^2
    >
    \frac{L_U}{2},
    \label{eq:condat_condition}
\end{equation}

which is verified during each loading update. The values of $\eta$ and
$\gamma$ used in the experiments are reported in
Appendix~\ref{app:numerical_settings}.

\subsection{Convergence and numerical safeguards}
\label{app:convergence}

Calendar-SPCA monitors convergence at three levels: refinement of the score
matrix, solution of the structured loading subproblem, and stabilization of
the complete low-rank factorization.

\paragraph{Score refinement.}

For fixed $V$, successive sweeps update all active score vectors using
Eqs.~\eqref{eq:score_update}--\eqref{eq:score_update_normalized}. Score
refinement terminates when

\begin{equation}
    \frac{
        \|U_{\mathcal{A}}^{(s+1)}
        -
        U_{\mathcal{A}}^{(s)}\|_F
    }{
        \|U_{\mathcal{A}}^{(s)}\|_F
        +
        \varepsilon_{\mathrm{num}}
    }
    <
    \varepsilon_U,
    \label{eq:score_convergence}
\end{equation}

or when the prescribed number of score sweeps is completed.

\paragraph{Loading convergence.}

The accuracy of the primal--dual loading solve is progressively refined over
the outer iterations. At outer iteration $t$, its target tolerance is

\begin{equation}
    \varepsilon_t
    =
    \max
    \left\{
        \varepsilon_{\min},
        \varepsilon_0\rho^t
    \right\},
    \qquad
    0<\rho<1.
    \label{eq:inner_tolerance_schedule}
\end{equation}

Every $q$ primal--dual iterations, the relative loading change is evaluated as

\begin{equation}
    \delta_V
    =
    \frac{
        \|V^{(\ell)}-V^{(\ell-q)}\|_F
    }{
        \|V^{(\ell-q)}\|_F
        +
        \varepsilon_{\mathrm{num}}
    }.
    \label{eq:relative_loading_change}
\end{equation}

A complementary fixed-point residual measures consistency with the primal and
dual updates. Define

\begin{equation}
P(V,Y)
=
\operatorname{soft}
\left(
V-\tau[\nabla f(V)+D_G^\top Y],
\tau\lambda_1
\right)
\end{equation}

and

\begin{equation}
Q(V,Y)
=
\operatorname{clip}
\left(
Y+\sigma D_GV,
-\lambda_{\mathrm{TV}},
\lambda_{\mathrm{TV}}
\right).
\end{equation}

The normalized primal--dual residual is

\begin{equation}
r_{\mathrm{PD}}
=
\frac{
\sqrt{
\|V-P(V,Y)\|_F^2/\tau^2
+
\|Y-Q(V,Y)\|_F^2/\sigma^2
}
}{
1+\|V\|_F+\|Y\|_F
}.
\label{eq:primal_dual_residual}
\end{equation}

The loading subproblem is considered converged when

\begin{equation}
    \delta_V<\varepsilon_t,
    \qquad
    r_{\mathrm{PD}}<\varepsilon_t,
    \label{eq:inner_convergence}
\end{equation}

and the loading objective satisfies the descent safeguard. A numerical margin

\begin{equation}
    10^{-12}
    \left(
        1+
        |J_U(V_{\mathrm{in}})|
    \right)
\end{equation}

is included in this check to accommodate floating-point precision. At the
maximum inner-iteration budget, a candidate satisfying the same descent
condition is retained. The pre-update primal and dual states remain active
whenever the descent condition is unmet.

\paragraph{Outer convergence.}

Let

\[
J^{(t)}
=
J\!\left(U^{(t)},V^{(t)}\right)
\]

denote the complete Calendar-SPCA objective and

\[
F^{(t)}
=
U^{(t)}V^{(t)\top}
\]

the corresponding fitted representation. Outer stabilization is monitored
through the relative objective change

\begin{equation}
    \delta_J^{(t)}
    =
    \frac{
        |J^{(t)}-J^{(t-1)}|
    }{
        |J^{(t-1)}|
        +
        \varepsilon_{\mathrm{num}}
    },
    \label{eq:outer_objective_change}
\end{equation}

and the relative reconstruction change

\begin{equation}
    \delta_F^{(t)}
    =
    \frac{
        \|F^{(t)}-F^{(t-1)}\|_F
    }{
        \|F^{(t-1)}\|_F
        +
        \varepsilon_{\mathrm{num}}
    }.
    \label{eq:outer_reconstruction_change}
\end{equation}

The fitting procedure terminates when

\begin{equation}
    \delta_J^{(t)}<\varepsilon_J,
    \qquad
    \delta_F^{(t)}<\varepsilon_F,
    \label{eq:outer_convergence}
\end{equation}

the current loading subproblem has converged, and the active component set has
remained unchanged during the iteration.

The reconstruction difference in
Eq.~\eqref{eq:outer_reconstruction_change} is evaluated exactly from the
$K\times K$ Gram matrices of consecutive factorizations, avoiding explicit
construction of the full $N\times M$ reconstructed matrices.

The numerical values of all convergence tolerances, iteration budgets, and
check intervals are collected in
Appendix~\ref{app:numerical_settings}. Component degeneracy and recovery are
described next.

\subsection{Initialization and component recovery}
\label{app:initialization_recovery}

Calendar-SPCA is initialized from a randomized truncated singular-value
decomposition of the centered data matrix $X_c$. For nominal rank $K$ and
oversampling parameter $s$, a Gaussian matrix

\[
\Omega
\in
\mathbb{R}^{M\times(K+s)}
\]

is used to construct an approximate dominant left singular subspace. A small
number of power iterations refines this subspace, followed by a reduced
singular-value decomposition. The initial factors are arranged so that

\begin{equation}
    \|u_k^{(0)}\|_2=1,
\end{equation}

with the corresponding singular-value amplitudes assigned to the loading
vectors $v_k^{(0)}$. This initialization follows the same score--loading scale
convention used throughout the Calendar-SPCA factorization.

During optimization, the active components are monitored for numerical
degeneracy. Component $k$ is selected for recovery when

\begin{equation}
    \|v_k\|_2
    <
    \varepsilon_{\mathrm{load}}
\end{equation}

or when its unnormalized score update satisfies

\begin{equation}
    \|\widetilde u_k\|_2
    <
    \varepsilon_{\mathrm{score}}.
\end{equation}

Redundancy between active loading vectors is assessed through their absolute
cosine similarity,

\begin{equation}
    c_{ij}
    =
    \frac{
        |v_i^\top v_j|
    }{
        \|v_i\|_2\|v_j\|_2
    },
    \label{eq:redundancy_cosine}
\end{equation}

with values above $c_{\max}$ identifying near-duplicate loading directions.
The condition number of the active loading Gram matrix,

\begin{equation}
    \kappa
    \left(
        V_{\mathcal{A}}^\top V_{\mathcal{A}}
    \right),
\end{equation}

provides an additional diagnostic for near-dependence among several
components.

When recovery is triggered, the component with the smallest conditional
reconstruction contribution is selected from the affected set. A new loading
direction is then extracted from residual variation in the data. The
candidate direction is projected away from the remaining active loadings and
refined through a small number of residual power iterations. Its associated
score vector is normalized and the recovered amplitude is assigned to the
loading vector.

Each component is given a finite number of recovery attempts. Components that
sustain a stable recovered direction remain in the active set
$\mathcal{A}$, while the final active set determines the effective rank

\begin{equation}
    K_{\mathrm{eff}}
    =
    |\mathcal{A}|.
\end{equation}

For reproducible presentation, a deterministic sign convention is applied
after fitting. For each active loading $v_k$, define

\begin{equation}
    j_k^\star
    =
    \arg\max_j |v_{jk}|.
\end{equation}

The pair $(u_k,v_k)$ is oriented so that
$v_{j_k^\star,k}>0$. This transformation preserves the rank-one component
$u_kv_k^\top$ and provides consistent signs across stored solutions and
visualizations.

The recovery thresholds, redundancy criteria, power-iteration settings, and
maximum number of recovery attempts are reported in
Appendix~\ref{app:numerical_settings}.

\subsection{Numerical settings}
\label{app:numerical_settings}

The numerical controls were kept fixed across the experiments, with the
outer-iteration budget increased for the full-data regularization-selection
fits. Table~\ref{tab:numerical_settings} summarizes the settings governing
score refinement, primal--dual optimization, convergence, initialization, and
component recovery.

\begin{table}[!ht]
\centering
\caption{
Numerical settings used in the Calendar-SPCA experiments.
}
\label{tab:numerical_settings}
\small
\renewcommand{\arraystretch}{1.2}

\begin{tabular}{lll}
\toprule
\textbf{Setting}
&
\textbf{Symbol}
&
\textbf{Value}
\\
\midrule

Score refinement
&
sweeps / $\varepsilon_U$
&
$10$ / $10^{-7}$
\\

Outer iteration budget
&
--
&
$500$ (factorial); $1000$ (selection)
\\

Outer tolerances
&
$(\varepsilon_J,\varepsilon_F)$
&
$(10^{-6},10^{-6})$
\\

Numerical safeguard
&
$\varepsilon_{\mathrm{num}}$
&
$10^{-12}$
\\

Inner iteration budget
&
--
&
$50\,000$
\\

Inner check interval
&
$q$
&
$25$
\\

Inner tolerance schedule
&
$(\varepsilon_0,\varepsilon_{\min},\rho)$
&
$(10^{-3},10^{-6},0.75)$
\\

Primal--dual controls
&
$(\eta,\gamma)$
&
$(0.99,1.0)$
\\

Degeneracy thresholds
&
$(\varepsilon_{\mathrm{load}},
  \varepsilon_{\mathrm{score}})$
&
$(10^{-10},10^{-12})$
\\

Redundancy thresholds
&
$(c_{\max},\kappa_{\max})$
&
$(0.995,10^{12})$
\\

Maximum recovery attempts
&
--
&
$3$
\\

Residual power iterations
&
--
&
$5$
\\

SVD power iterations / oversampling
&
--
&
$2/5$
\\

\bottomrule
\end{tabular}
\end{table}


\section{Additional robustness and ablation}
\label{app:extended_analysis}

This appendix complements the main factorial analysis with three additional
views of Calendar-SPCA behaviour. The first summarizes the numerical
reliability of the complete factorial experiment. The second isolates the
individual contributions of loading sparsity and calendar coherence through a
matched penalty ablation. The third examines the dependence of the objective
balance on sample size.

\subsection{Reliability of the factorial experiment}
\label{app:convergence_diagnostics}

The factorial design comprised $5{,}880$ Calendar-SPCA fits spanning the full
grid of regularization strengths, sample sizes, nominal ranks, and
repetitions. Among these fits, $3{,}499$ ($59.5\%$) satisfied the complete
convergence criterion defined in Appendix~\ref{app:convergence}, and $2{,}381$
completed the prescribed budget of 500 outer iterations.

Table~\ref{tab:convergence_diagnostics} summarizes the final numerical
diagnostics of both groups.

\begin{table}[!ht]
\centering
\caption{
Numerical diagnostics of the Calendar-SPCA factorial experiment.
Outer iteration counts and final relative changes are reported as medians;
$K_{\mathrm{eff}}/K$ is averaged across fits.
}
\label{tab:convergence_diagnostics}
\small
\renewcommand{\arraystretch}{1.15}

\begin{tabular}{lrrrrrr}
\toprule
Status
& Fits
& Outer iter.
& $\Delta J_{\mathrm{rel}}$
& $\Delta F_{\mathrm{rel}}$
& $K_{\mathrm{eff}}/K$
& Full rank (\%) \\
\midrule

Converged
& 3499
& 191
& $5.87\times10^{-12}$
& $9.73\times10^{-7}$
& 0.931
& 84.1 \\

Iteration-budget
& 2381
& 500
& $8.73\times10^{-8}$
& $5.73\times10^{-5}$
& 0.999
& 99.6 \\

\bottomrule
\end{tabular}
\end{table}

The converged fits show very small final objective and reconstruction changes,
consistent with the stopping tolerances used throughout the study. Fits
reaching the iteration budget also retain nearly all requested components and
typically exhibit small final changes, indicating advanced numerical
stabilization within the fixed computational budget.

The factorial summaries reported in Section~\ref{sec:results_regularization}
use a reliability criterion based on repeated convergence. A configuration is
included when at least four of its five repetitions satisfy the complete
convergence criterion. This rule ensures that each aggregated grid value is
supported by consistent repeated solutions and provides the reliability mask
used in the regularization landscape.

\subsection{Isolating sparsity and calendar coherence}
\label{app:penalty_ablation}

The regularization landscape in
Section~\ref{sec:results_regularization} shows the joint behaviour of the
two Calendar-SPCA penalties over the complete factorial grid. A complementary
view is obtained by comparing a small set of representative configurations
that isolate the individual and combined effects of loading sparsity and
calendar coherence.

To make these effects directly visible, we consider the four configurations

\[
(\lambda_1,\lambda_{\mathrm{TV}})
\in
\left\{
(0,0),\,
(3,0),\,
(0,3),\,
(3,3)
\right\}.
\]

The value $3$ provides a clearly regularized regime in which the effect of
each penalty can be read directly from the fitted loading structure.
Table~\ref{tab:penalty_ablation} reports the corresponding averages over the
nine reliable $(N,K)$ scenarios common to all four configurations.

\begin{table}[!ht]
\centering
\caption{
Matched comparison isolating the effects of the sparsity and
calendar-coherence penalties. Values are averaged over the nine reliable
$(N,K)$ scenarios common to all four configurations.
}
\label{tab:penalty_ablation}
\small
\renewcommand{\arraystretch}{1.15}

\begin{tabular}{ccrrrr}
\toprule
$\lambda_1$
& $\lambda_{\mathrm{TV}}$
& EV
& Sparsity
& RTV
& Regions \\
\midrule

0 & 0 & 0.387 & 0.000 & 0.779 & 1.00 \\
3 & 0 & 0.346 & 0.691 & 0.910 & 7.35 \\
0 & 3 & 0.360 & 0.000 & 0.204 & 1.00 \\
3 & 3 & 0.310 & 0.657 & 0.386 & 3.64 \\

\bottomrule
\end{tabular}
\end{table}

The comparison shows a clear division of roles between the two penalties.
When $\lambda_{\mathrm{TV}}=0$, increasing $\lambda_1$ from 0 to 3 raises
mean loading sparsity from 0 to 0.691. The $\ell_1$ term therefore acts as
the main mechanism controlling support size and temporal selectivity.

When $\lambda_1=0$, increasing $\lambda_{\mathrm{TV}}$ from 0 to 3 reduces
the mean relative total variation from 0.779 to 0.204 while preserving dense
supports. The graph-TV term therefore acts primarily on the internal
organization of the loading field, promoting coherent values across adjacent
calendar positions.

When both penalties are active, these two effects combine within the same
representation. The resulting loadings remain strongly sparse
(mean sparsity $=0.657$) and substantially more coherent than under sparsity
alone (RTV $=0.386$ versus $0.910$). The mean number of effective regions
also decreases from 7.35 to 3.64, showing that calendar coherence organizes
the selected support into a smaller number of dominant temporal structures.

This matched comparison supports the interpretation adopted throughout the
paper. The $\ell_1$ penalty determines how much of the calendar domain is
active, while graph total variation shapes how that active support is
connected and organized. Their combination yields the sparse and coherent
calendar patterns that characterize Calendar-SPCA.

\subsection{Sample-size dependence of the objective balance}
\label{app:objective_scaling}

The factorial experiment also provides information about how the relative
weight of the regularization terms evolves with sample size. This effect was
examined at the common configuration
$K=15$ and
$\lambda_1=\lambda_{\mathrm{TV}}=1$ for

\[
N\in\{5{,}000,\;10{,}000,\;15{,}000,\;20{,}000\}.
\]

For each fitted model, the Calendar-SPCA objective was decomposed as

\[
J
=
R+P_1+P_{\mathrm{TV}},
\]

with

\[
R
=
\frac{1}{2}
\|X_c-UV^\top\|_F^2,
\qquad
P_1
=
\lambda_1\|V\|_{1,1},
\qquad
P_{\mathrm{TV}}
=
\lambda_{\mathrm{TV}}\|D_GV\|_{1,1}.
\]

Table~\ref{tab:objective_scaling} reports the two penalty terms relative to
the reconstruction contribution. Values are averaged over the five converged
repetitions available at each sample size.

\begin{table}[!ht]
\centering
\caption{
Relative objective balance as a function of sample size for
$K=15$ and $\lambda_1=\lambda_{\mathrm{TV}}=1$.
}
\label{tab:objective_scaling}
\small
\renewcommand{\arraystretch}{1.15}

\begin{tabular}{rrrr}
\toprule
$N$
& $P_1/R$
& $P_{\mathrm{TV}}/R$
& $(P_1+P_{\mathrm{TV}})/R$
\\
\midrule

5\,000
& 0.105
& 0.052
& 0.157
\\

10\,000
& 0.081
& 0.042
& 0.124
\\

15\,000
& 0.070
& 0.037
& 0.107
\\

20\,000
& 0.062
& 0.033
& 0.096
\\

\bottomrule
\end{tabular}
\end{table}

The relative contribution of both regularizers decreases systematically as
the number of observations increases. The combined
penalty-to-reconstruction ratio changes from $0.157$ at $N=5{,}000$ to
$0.096$ at $N=20{,}000$, with the same progression visible independently for
the sparsity and calendar-TV terms.

This behaviour shows that a fixed numerical pair
$(\lambda_1,\lambda_{\mathrm{TV}})$ corresponds to a sample-size-dependent
balance between reconstruction and regularization. The observation provides
additional context for the sparsity trends in
Section~\ref{sec:results_robustness} and motivates future investigation of
sample-size-aware regularization scaling.


\section{Baseline implementation details}
\label{app:baselines}

This appendix documents the reference methods used in the cross-method
comparison of Section~\ref{sec:results_representation_quality}. It specifies
the fitted baseline formulations, their implementation, and the numerical
procedures used for regularization selection.

\subsection{Reference methods}
\label{app:baseline_methods}

All reference methods were evaluated at the common rank $K=15$ using the same
normalized annual profiles and centering convention as Calendar-SPCA.

\paragraph{PCA.}

Ordinary PCA was fitted directly to the complete centered matrices and provides
the rank-matched linear reconstruction reference used throughout the
comparison.

\paragraph{Sparse PCA.}

Sparse PCA was represented by a high-dimensional, matrix-free implementation
of the alternating \texttt{arrayspc} scheme of
Zou et al.~\cite{zou2006sparse}. The method estimates sparse loading
directions through iterative singular-vector updates and soft-thresholding.

For scalability on the complete smart-meter matrices, products involving the
sample covariance operator were evaluated without explicitly forming
$X_c^\top X_c$, using

\begin{equation}
    X_c^\top(X_cB).
\end{equation}

The dominant rank-$15$ initialization was obtained through truncated SVD. The
matrix-free implementation was validated against
\texttt{elasticnet::arrayspc} on a tractable problem, reproducing the fitted
loading vectors and adjusted explained variance to numerical precision.

\paragraph{SPCA-TV.}

SPCA-TV was fitted using the \texttt{PCAL1L2TV} implementation associated
with de Pierrefeu et al.~\cite{depierrefeu2018spcatv}. The method extracts
successive rank-one components under $\ell_1$, quadratic, and total-variation
regularization.

For the present comparison, its structural operator was constructed from the
same cyclic calendar neighbourhood used by Calendar-SPCA. SPCA-TV applies its
native isotropic total variation across the directional calendar operators.
The common relative total-variation diagnostic reported in
Section~\ref{sec:results_representation_quality} was subsequently evaluated
for all sparse methods using the anisotropic graph definition of
Eq.~\eqref{eq:relative_tv} on the same calendar graph.

The regularization parameters of SPCA and SPCA-TV were selected through the
method-specific procedures described next.

\subsection{Regularization selection}
\label{app:baseline_selection}

Regularization parameters for SPCA and SPCA-TV were selected numerically before
inspection of the fitted component maps. Both methods used one-dimensional
regularization paths and the same L-curve principle adopted for the final
Calendar-SPCA configurations.

\paragraph{Sparse PCA.}

For SPCA, the soft-thresholding parameter was expressed relative to the
data-dependent reference scale

\begin{equation}
    \tau_{\mathrm{ref}}
    =
    \max_k
    \left\{
        \sigma_k^2
        \|v_k^{\mathrm{PCA}}\|_\infty
    \right\},
\end{equation}

where $\sigma_k$ and $v_k^{\mathrm{PCA}}$ are obtained from the rank-$15$
PCA initialization. The \texttt{arrayspc} regularization parameter was then
written as

\begin{equation}
    \mathrm{para}
    =
    \gamma\tau_{\mathrm{ref}}.
\end{equation}

The candidate path was

\[
\begin{aligned}
\Gamma = \{&
0.0025,\;0.0035,\;0.005,\;0.0075,\;0.01,\;0.0125,\;0.015,\;0.0175,\\
&0.02,\;0.025,\;0.03,\;0.035,\;0.04,\;0.05,\;0.06,\;0.08
\}.
\end{aligned}
\]

For each value of $\gamma$, the L-curve was formed from the loading
$\ell_1$ magnitude and reconstruction residual. Maximum discrete curvature
selected

\begin{equation}
    \gamma^*
    =
    0.02
    \qquad
    \text{for GoiEner},
\end{equation}

and

\begin{equation}
    \gamma^*
    =
    0.05
    \qquad
    \text{for Low Carbon London}.
\end{equation}

\paragraph{SPCA-TV.}

The three SPCA-TV penalties were parameterized through a common multiplicative
factor $\alpha$ relative to the data-dependent scale
$\lambda_{1,\max}$ used by the implementation,

\begin{equation}
    (\lambda_1,\lambda_2,\lambda_{\mathrm{TV}})
    =
    \alpha\lambda_{1,\max}
    (0.05,\;1,\;0.10).
\end{equation}

The candidate path was

\[
\mathcal{A}_{\alpha}
=
\{
0.1,\;0.2,\;0.35,\;0.5,\;0.75,\;1,\;1.5,\;2,\;3,\;5,\;7,\;10
\}.
\]

The L-curve combined the native regularization magnitude with reconstruction
residual, and maximum discrete curvature selected

\begin{equation}
    \alpha^*
    =
    1.5
    \qquad
    \text{for GoiEner},
\end{equation}

and

\begin{equation}
    \alpha^*
    =
    0.5
    \qquad
    \text{for Low Carbon London}.
\end{equation}

For GoiEner, the SPCA and SPCA-TV regularization paths were evaluated on the
same fixed deterministic sample of $N=5{,}000$ profiles. The selected
dimensionless factors were subsequently transferred to fits on the complete
$N=87{,}187$ dataset, with their data-dependent regularization scales
recomputed from the full centered matrix. For Low Carbon London, both
regularization paths were evaluated directly on the complete
$N=5{,}359$ dataset.

All selected full-data representations retained the complete rank $K=15$.
The resulting reconstruction and structural metrics are summarized in the
following section.

\subsection{Final cross-method summary}
\label{app:baseline_summary}

Table~\ref{tab:baseline_summary} reports the final rank-matched comparison
obtained with the selected configurations of PCA, SPCA, SPCA-TV, and
Calendar-SPCA. All methods were evaluated at $K=15$ on the complete datasets.

\begin{table}[!ht]
\centering
\caption{
Final cross-method comparison at $K=15$. EV denotes explained variance,
PCA retention expresses EV relative to the corresponding PCA solution,
sparsity is the mean proportion of loading coefficients at or below the
support threshold, and RTV is the mean relative total variation over the
common cyclic calendar graph.
}
\label{tab:baseline_summary}
\small
\renewcommand{\arraystretch}{1.15}

\begin{tabular}{llrrrr}
\toprule
\textbf{Dataset}
&
\textbf{Method}
&
\textbf{EV (\%)}
&
\textbf{PCA retention (\%)}
&
\textbf{Sparsity (\%)}
&
\textbf{RTV}
\\
\midrule

GoiEner
& PCA
& 41.95
& 100.00
& 0.00
& 0.865
\\

& SPCA
& 41.29
& 98.44
& 47.12
& 0.994
\\

& SPCA-TV
& 34.39
& 81.98
& 42.10
& 0.792
\\

& Calendar-SPCA
& 40.66
& 96.92
& 61.95
& 0.615
\\

\midrule

LCL
& PCA
& 27.63
& 100.00
& 0.00
& 0.858
\\

& SPCA
& 26.14
& 94.63
& 64.24
& 1.052
\\

& SPCA-TV
& 25.76
& 93.22
& 19.02
& 0.509
\\

& Calendar-SPCA
& 22.90
& 82.90
& 81.50
& 0.450
\\

\bottomrule
\end{tabular}
\end{table}

The final comparison shows the distinct structural profiles of the four
representations. PCA provides the rank-matched reconstruction reference,
SPCA introduces substantial loading sparsity while preserving a high fraction
of PCA variance, and SPCA-TV emphasizes local regularity over the calendar
neighbourhood. Calendar-SPCA combines high loading selectivity with strong
calendar coherence while retaining substantial explained variance on both
datasets.
These numerical results complement the cross-method analysis in
Section~\ref{sec:results_representation_quality} and the Calendar-SPCA
component representations in
Section~\ref{sec:results_interpretable_components}.

\end{document}